\documentclass[11pt]{article}

\usepackage[preprint]{acl}
\usepackage{times}
\usepackage{latexsym}
\usepackage{float}
\usepackage[T1]{fontenc}
\usepackage[utf8]{inputenc}

\usepackage{microtype}

\usepackage{inconsolata}

\usepackage{graphicx}

\usepackage{longtable}
\usepackage{amssymb}
\usepackage{tabularx,makecell,multirow}
\usepackage{booktabs}
\usepackage{enumitem}
\usepackage[most]{tcolorbox}
\usepackage{listings}
\usepackage{listingsutf8}
\usepackage{xspace}
\usepackage{cleveref}
\usepackage{stfloats}
\usepackage{mathtools}

\usepackage{atbegshi}

\newtcolorbox{promptbox}[1][Prompt]{
    colback=gray!5,
    colframe=orange!80!black,
    fonttitle=\bfseries,
    title=#1,
    rounded corners,
    boxrule=1.5pt,
    left=4mm,
    right=4mm,
    top=3mm,
    bottom=3mm,
    breakable  % This allows the box to span multiple pages
}

\lstdefinestyle{promptcode}{
    basicstyle=\footnotesize\ttfamily,
    breaklines=true,
    backgroundcolor=\color{gray!3},
    frame=single,
    framesep=3pt,
    xleftmargin=10pt,
    xrightmargin=10pt
}

\newcommand{\AlgoName}{\texttt{DREAM}\xspace}
\newcommand{\mycomment}[3]{}

\newcommand{\CF}{\text{CF}\xspace}

\newcommand{\ClaimSupport}{\text{supp}\xspace}
\newcommand{\ClaimPartSupport}{\text{part}\xspace}
\newcommand{\ClaimContradict}{\text{con}\xspace}

\usepackage{colortbl}
\usepackage{xcolor}
\usepackage{soul}
\definecolor{forestgreen}{RGB}{34, 139, 34}

\definecolor{writingcolor}{RGB}{232, 245, 255}    % soft blue
\definecolor{aligncolor}{RGB}{232, 255, 238}      % soft green
\definecolor{analysiscolor}{RGB}{255, 243, 224}   % soft orange
\definecolor{factcolor}{RGB}{243, 232, 255}       % soft purple

\newcommand{\writinghl}[1]{{\sethlcolor{writingcolor}\hl{#1}}}
\newcommand{\alignhl}[1]{{\sethlcolor{aligncolor}\hl{#1}}}
\newcommand{\analysishl}[1]{{\sethlcolor{analysiscolor}\hl{#1}}}
\newcommand{\facthl}[1]{{\sethlcolor{factcolor}\hl{#1}}}

\title{
\begin{tikzpicture}[remember picture, overlay]
    \node[anchor=east, xshift=-0.2cm, yshift=0.3cm] at (0,0) {\includegraphics[height=1.7cm]{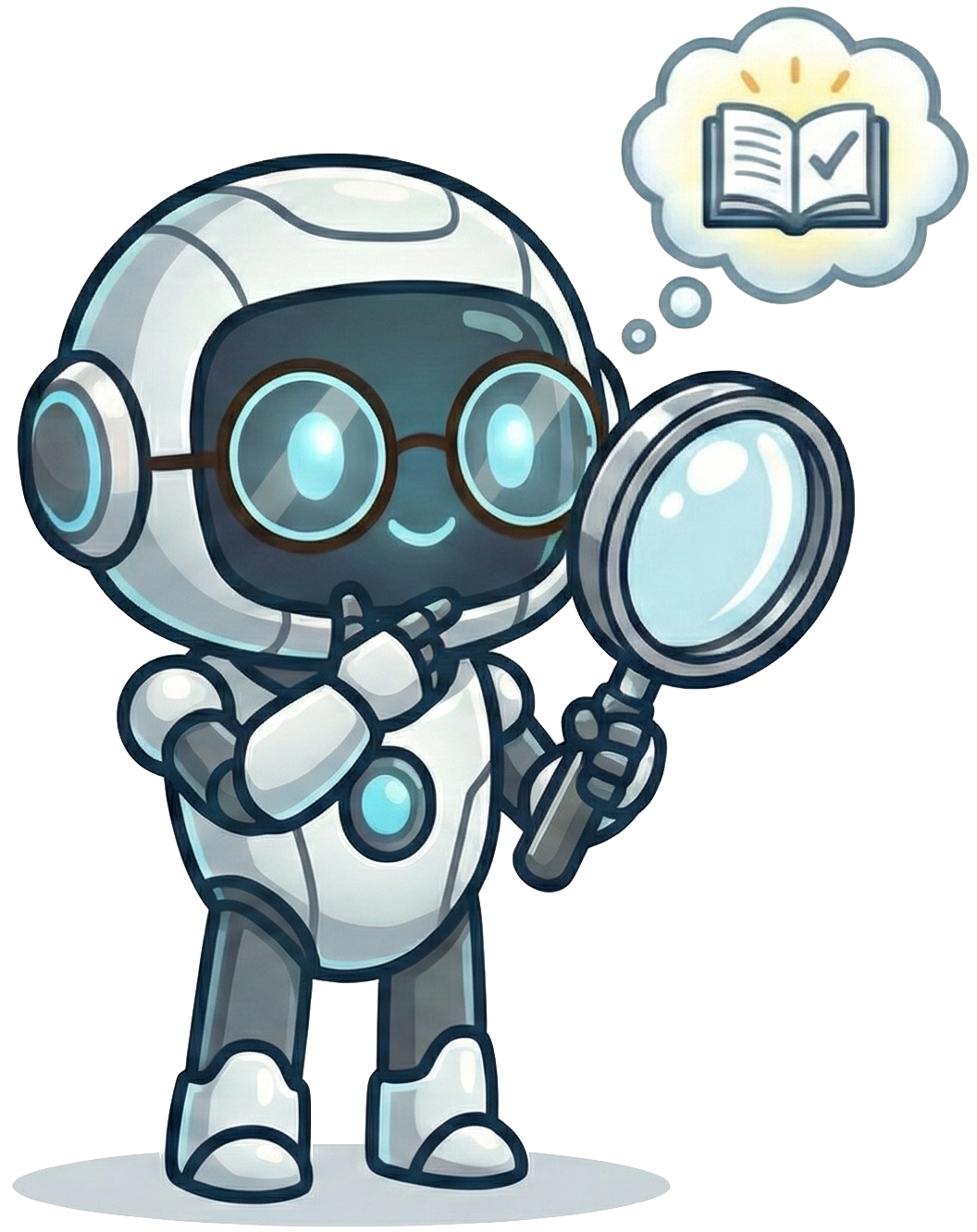}};
  \end{tikzpicture}%
DREAM: Deep Research Evaluation with Agentic Metrics
}

\author{
  \textbf{Elad Ben Avraham\textsuperscript{1}%
    \thanks{Equal contribution}%
    \thanks{Correspondence to: \href{mailto:eladba@amazon.com}{eladba@amazon.com}}} \quad
  \textbf{Changhao Li\textsuperscript{2}\footnotemark[1]%
    \thanks{Work done during an internship at AWS Agentic AI}} \quad
  \textbf{Ron Dorfman\textsuperscript{1}\footnotemark[1]} \quad
  \textbf{Roy Ganz\textsuperscript{1}}
\\
  \textbf{Oren Nuriel\textsuperscript{1}} \quad
  \textbf{Amir Dudai\textsuperscript{1}} \quad
  \textbf{Aviad Aberdam\textsuperscript{1}} \quad
  \textbf{Noah Flynn\textsuperscript{1}}
\\
  \textbf{Elman Mansimov\textsuperscript{1}} \quad 
  \textbf{Adi Kalyanpur\textsuperscript{1}} \quad
  \textbf{Ron Litman\textsuperscript{1}}
\\[0.1em]
  \textsuperscript{1}AWS Agentic AI \quad
  \textsuperscript{2}Georgia Institute of Technology
}

\begin{document}
\maketitle

\begin{abstract}
Deep Research Agents generate analyst-grade reports, yet evaluating them remains challenging due to the absence of a single ground truth and the multidimensional nature of research quality. Recent benchmarks propose distinct methodologies, yet they suffer from the \emph{Mirage of Synthesis}, 
where strong surface-level fluency and citation alignment can obscure underlying factual and reasoning defects. We characterize this gap by introducing a taxonomy across four verticals that exposes a critical \emph{capability mismatch}: static evaluators inherently lack the tool-use capabilities required to assess temporal validity and factual correctness. To address this, we propose \textbf{\AlgoName{}} (\underline{D}eep \underline{R}esearch \underline{E}valuation with \underline{A}gentic \underline{M}etrics), a framework that instantiates the principle of \emph{capability parity} by making evaluation itself agentic. \AlgoName{} structures assessment through an evaluation protocol combining query-agnostic metrics with adaptive metrics generated by a tool-calling agent, enabling temporally aware coverage, grounded verification, and systematic reasoning probes. Controlled evaluations demonstrate \AlgoName{} is significantly more sensitive to factual and temporal decay than existing benchmarks, offering a scalable, reference-free evaluation paradigm.
\end{abstract}
\section{Introduction}
% \begin{figure}[t!]
%     \centering
%     % \includegraphics[trim=4.8cm 4.54cm 4.5cm 4.54cm,width=\linewidth]{figures/ErrorInject_KnowledgeCutoff.pdf}
%     % \includegraphics[trim=4.8cm 4.54cm 4.5cm 4.54cm,width=.97\linewidth]{figures/teaser-new-vertical.pdf}
%     \includegraphics[trim=4.8cm 4.54cm 4.5cm 4.54cm,width=.8\linewidth]{figures/teaser-new-vertical-flipped.pdf}
%     % \caption{\textbf{Capturing Overlooked Dimensions of Research Quality.} Left: \AlgoName{} detects factual errors injected in a controlled experiment. Right: \AlgoName{} captures time-sensitive validity gaps, penalizing outdated reports.}
%     \label{fig:error_inject_knowledge_cutoff}
%     \caption{\textbf{Capturing Overlooked Dimensions of Research Quality.} \textbf{Top:} \AlgoName{} captures time-sensitive validity gaps, penalizing outdated reports. \textbf{Bottom:} \AlgoName{} detects factual errors injected in a controlled experiment.}
%     \label{fig:error_inject_knowledge_cutoff}
%     \vspace{-2em}
% \end{figure}

\begin{figure*}[t]
    \centering
    \includegraphics[trim=4.8cm 4.54cm 4.5cm 4.54cm,width=\linewidth]{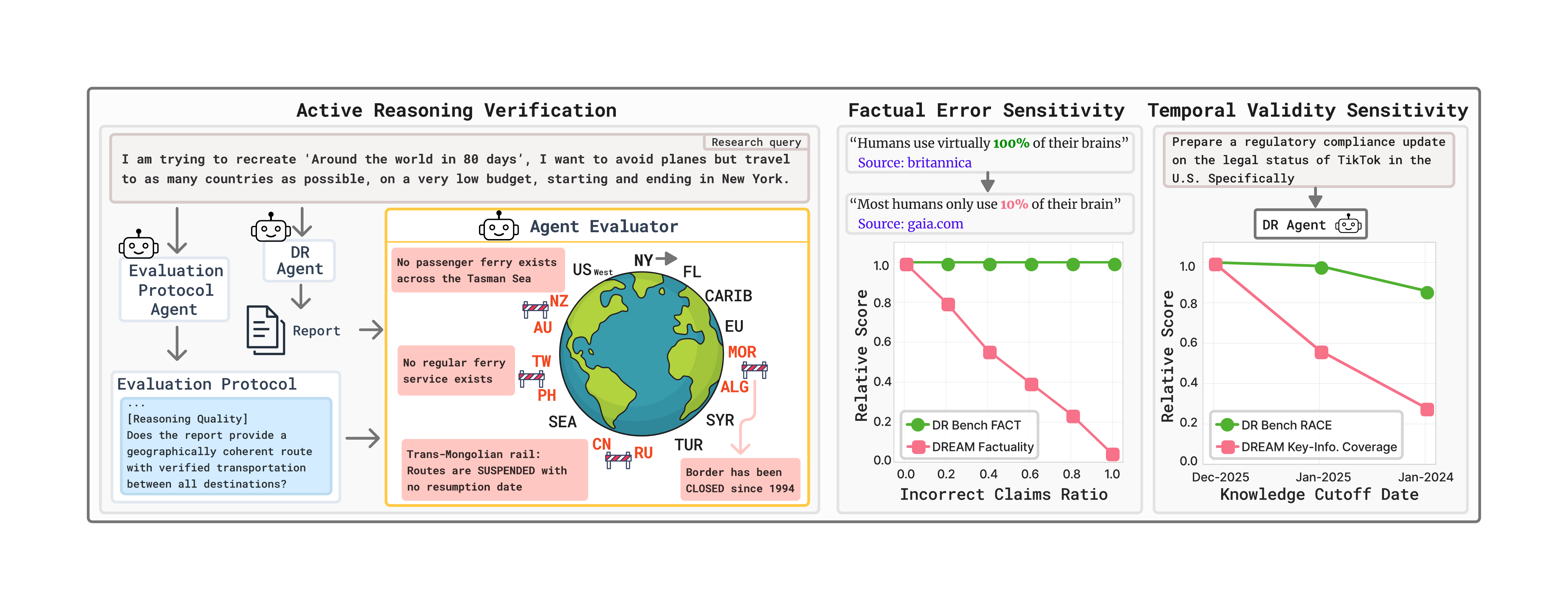}
    % \caption{\textbf{Capturing Overlooked Dimensions of Research Quality.} Left: \AlgoName{} detects factual errors injected in a controlled experiment. Right: \AlgoName{} captures time-sensitive validity gaps, penalizing outdated reports. \textcolor{red}{edit}}
    \caption{\textbf{Capturing Overlooked Dimensions of Research Quality.} \AlgoName{} actively verifies the reasoning of generated reports by probing external sources (left), detects factual errors injected in a controlled experiment (middle), and captures time-sensitive validity gaps by penalizing outdated reports (right).}
    \label{fig:teaser}
    \vspace{-2ex}
\end{figure*}

Large Language Models (LLMs) increasingly support autonomous, tool-using agents that perform complex, open-ended tasks. Among these, \emph{Deep Research Agents} (DRAs) have emerged as a dominant paradigm. Given a broad query, they retrieve information from external sources, synthesize evidence, and produce long-form research reports. Such reports are inherently open-ended; unlike traditional question answering, where correctness can often be verified against a singular ground truth, deep research admits multiple valid trajectories. For a given query, two experts will inevitably produce distinct reports of potentially equal quality. This variability makes the assessment of deep research fundamentally complex, requiring a transition from single-answer correctness toward a high-dimensional evaluation of report quality.

Recognizing this challenge, the research community has shown growing interest in Deep Research Evaluation (DRE), with several recent works proposing benchmarks and datasets to assess DRAs \cite{coelho2025deepresearchgym, du2025deepresearch, patel2025deepscholar, sharma2025researchrubrics, wan2025deepresearch, wang2025liveresearchbench}. However, a critical examination of these frameworks reveals a systematic limitation. While current approaches effectively assess surface-level dimensions, such as writing fluency and citation alignment, they remain largely insensitive to failures in \emph{factual correctness}, \emph{temporal validity}, and \emph{substantive reasoning}. Therefore, fluent, well-cited reports can receive high scores despite containing obsolete information or flawed logic. We term this the \emph{Mirage of Synthesis} -- an illusion of quality created by surface-level coherence despite underlying factual and reasoning flaws.

To characterize this observation and unify the evaluation landscape, we propose a taxonomy that organizes DRE into four fundamental verticals: \emph{Presentation Quality}, \emph{Task Compliance}, \emph{Analytical Depth}, and \emph{Source Quality}. By mapping the metrics of existing benchmarks into this taxonomy, we trace their limitations to the fundamental design choices. Current benchmarks typically rely on either human-curated rubrics, which are highly reliable but prohibitively expensive, dataset-specific, and prone to becoming outdated, or LLM-as-a-judge paradigms limited to the LLM’s static internal knowledge. These approaches and their associated citation-verification workflows operate as \emph{static observers}, i.e., they lack access to external tools, temporal awareness, and the ability to independently gather and verify evidence. Our taxonomy thus reveals a unifying diagnosis: a capability mismatch, where evaluators lack the abilities required to assess the dimensions they purport to measure. This motivates the principle of \textbf{capability parity} -- the evaluator should possess a similar set of capabilities as the researcher, including the ability to retrieve, verify, and reason over information.

We instantiate this principle in \underline{D}eep \underline{R}esearch \underline{E}valuation with \underline{A}gentic \underline{M}etrics (\textbf{\AlgoName}), a framework that makes evaluation itself agentic. \AlgoName{} structures assessment through an \emph{evaluation protocol} that combines query-agnostic static metrics with query-adaptive ones constructed by a tool-calling agent. By independently researching the query and cross-referencing external evidence, \AlgoName{} provides temporally aware fact-checking and substantive depth assessment. 
% This behavior is illustrated in \Cref{fig:teaser}: the left panel demonstrates \AlgoName{}'s active reasoning probes against external sources; moreover, as shown in the middle and right panels, \AlgoName{}'s scores degrade appropriately as factual errors increase and report knowledge becomes outdated, while existing benchmarks remain largely insensitive to both.
This behavior is illustrated in \Cref{fig:teaser}: the left panel depicts \AlgoName{}'s active reasoning verification against external sources, while the middle and right panels demonstrate that \AlgoName{}'s scores degrade appropriately as factual errors increase and report knowledge becomes outdated, whereas existing benchmarks remain largely insensitive to both.

% the left panel depicts our active probing mechanism, while the middle and right panels demonstrate that \AlgoName{}'s scores degrade appropriately as factual errors increase and knowledge becomes outdated—sensitivities that existing benchmarks lack.

In summary, our main contributions are:
\vspace{-1ex}
\begin{itemize}[leftmargin=*, itemsep=0.1pt]
    \item We examine existing DRE benchmarks to identify critical gaps and the ``Mirage of Synthesis'', a systemic failure where they lack the capacity to detect factual, temporal, and logical degradation.
    \item We propose a taxonomy to unify DRE into four key verticals and identify a capability mismatch in current benchmarks, where evaluators lack the active retrieval tools and reasoning required for independent, temporally-aware verification.
    % \item We propose a taxonomy to unify DRE into four key verticals and trace limitations to design choices, identifying a capability mismatch where human judges are costly, dataset-dependent, and prone to obsolescence, while standard LLM-as-a-Judge evaluators lack the necessary tools for verification.
    \item We introduce \AlgoName{}, an agentic evaluation framework grounded in the principle of capability parity, replacing passive scoring with active verification via a two-phase Protocol Creation and Execution workflow.
    \item We empirically validate through controlled experiments a suite of agentic metrics---\emph{Key-Information Coverage}, \emph{Reasoning Quality}, and \emph{Factuality}--showing that \AlgoName{} is substantially more sensitive to temporal degradation and extrinsic factual errors than existing benchmarks.
\end{itemize}

\section{Deep Research Evaluation Landscape}
\label{sec:taxonomy}

% Evaluating automatically generated research reports is inherently complex, as report quality depends on multiple dimensions that collectively determine usefulness and trustworthiness. 

Existing DRE benchmarks propose a variety of metrics to capture the multidimensional nature of research quality, yet they differ substantially in terminology, scope, and implementation. This fragmentation makes it difficult to reason about which aspects of deep research are well evaluated, which remain under-evaluated, and why certain failure modes persist across benchmarks. To address this, we introduce a unifying taxonomy that organizes existing criteria into a common structure and use it to analyze the current evaluation landscape.

\subsection{A Unifying Taxonomy}\label{subsec:taxonomy}

To systematically derive the taxonomy, we employed an agentic pipeline that processed the evaluation metrics from the benchmarks summarized in \Cref{tab:taxonomy}. The pipeline extracted granular, leaf-level criteria from benchmark source code and documentation, embedded them semantically, and clustered them into coherent evaluation dimensions. This data-driven approach ensures the taxonomy is grounded in the actual implementation details of existing metrics, providing a structured, bottom-up synthesis of the landscape. Full details of this derivation process are provided in Appendix~\ref{app:taxonomy_pipeline}.

\newcommand{\writingcell}{\cellcolor{writingcolor}\raggedright Presentation Quality}
\newcommand{\aligncell}{\cellcolor{aligncolor}\raggedright Task Compliance}
\newcommand{\analysiscell}{\cellcolor{analysiscolor}\raggedright Analytical Depth}
\newcommand{\factcell}{\cellcolor{factcolor}\raggedright Source Quality}

\begin{table*}[t]
\centering
\caption{\textbf{Taxonomic breakdown of deep research evaluation benchmarks.} Metrics from existing benchmarks mapped to our four-dimensional taxonomy, alongside their creation and execution methodologies.}
\vspace{-1ex}
\small
\begin{tabularx}{\textwidth}{lllcc}
\toprule
\textbf{Benchmark} & \textbf{Vertical} & \textbf{Metric Name} & \textbf{Creation} & \textbf{Execution} \\
\midrule
DeepResearchGym & \writingcell    & Clarity                           & \textcolor{gray}{N/A} & LLM \\
\cite{coelho2025deepresearchgym} & \aligncell      & Report Relevance                  & LLM + Human                   & LLM \\
                & \analysiscell   & Insightfulness                    & \textcolor{gray}{N/A} & LLM \\
                & \factcell       & Retrieval Faithfulness            & \textcolor{gray}{N/A} & Workflow \\
\midrule
DeepResearch Bench & \writingcell    & RACE (Presentation Quality)                   & LLM  & LLM \\
\cite{du2025deepresearch} & \aligncell      & RACE (Inst., Comp.)            & LLM  & LLM \\
                & \analysiscell   & RACE (Insight)                    & LLM  & LLM \\
                & \factcell       & FACT                              & \textcolor{gray}{N/A} & Workflow \\
\midrule
ResearchRubrics & \writingcell    & Communication Quality             & Human & LLM \\
\cite{sharma2025researchrubrics} & \aligncell      & Explicit/Implicit Req., Inst. Follow.         & Human & LLM \\
                & \analysiscell   & Synthesis of Information          & Human & LLM \\
                & \factcell       & Use of References                 & Human & LLM \\
\midrule
DeepResearch Arena & \writingcell    & ACE                            & LLM  & LLM \\
\cite{wan2025deepresearch} & \aligncell      & ACE                            & LLM  & LLM \\
                & \analysiscell   & ACE                               & LLM  & LLM \\
                & \factcell       & KAE                               & \textcolor{gray}{N/A} & Workflow \\
\midrule
LiveResearchBench & \writingcell    & Presentation \& Organization   & LLM + Human & LLM \\
\cite{wang2025liveresearchbench} & \aligncell      & Coverage \& Comprehensiveness  & LLM + Human & LLM \\
                & \analysiscell   & Analysis Depth                    & \textcolor{gray}{N/A} & LLM \\
                & \factcell       & Factual Cons., Citation Assoc./Acc.              & \textcolor{gray}{N/A} & LLM / Workflow \\
\midrule
\AlgoName{} (Ours) & \writingcell    & Writing Quality                & \textcolor{gray}{N/A} & LLM \\
                & \aligncell      & Key-Information Coverage          & Agent & LLM \\
                & \analysiscell   & Reasoning Quality                 & Agent & Agent \\
                & \factcell       & Factuality, Citation Integrity & \textcolor{gray}{N/A} & Workflow \\
\bottomrule
\end{tabularx}
\label{tab:taxonomy}
\vspace{-2ex}
\end{table*}

\vspace{0.4ex}
\noindent \writinghl{\textbf{Presentation Quality}}
% Presentation Quality
% Evaluates how effectively a report communicates its findings. This includes stylistic aspects such as clarity of expression and sentence fluency, as well as structural properties such as section organization, and appropriate use of headings. Metrics in this vertical assess whether information is conveyed clearly and professionally, independent of the correctness or depth of the content.
evaluates how effectively a report communicates its findings. This includes stylistic aspects such as clarity of expression and sentence fluency, as well as structural properties such as section organization and appropriate use of headings. Metrics in this vertical assess whether information is conveyed clearly and professionally, independent of content correctness or depth.

\vspace{0.4ex}
\noindent \alignhl{\textbf{Task Compliance}}
%Task Compliance
% Assesses whether a report fulfills the requirements of the research query. This includes explicit constraints (\textit{e.g.}, format, scope, or requested comparisons) as well as implicit expectations of comprehensive coverage. Metrics in this vertical evaluate breadth of information, recall of key points, and whether the report addresses aspects that an informed expert would reasonably expect.
assesses whether a report fulfills the requirements of the research query. This includes explicit constraints (\textit{e.g.}, format, scope, or requested comparisons) as well as implicit expectations of comprehensive coverage. Metrics in this vertical evaluate coverage breadth, recall of key information, and adherence to the query.

\vspace{0.4ex}
\noindent \analysishl{\textbf{Analytical Depth}}
%Analytical Depth
captures the intellectual rigor of the report’s synthesis. This includes the quality of logical reasoning, causal explanations, and critical evaluation of evidence. High analytical depth reflects coherent reasoning that goes beyond aggregating retrieved facts, demonstrating structured argumentation and meaningful synthesis.

\vspace{0.4ex}
\noindent \facthl{\textbf{Source Quality}}
%Source Quality
measures the reliability of the evidence supporting a report’s claims. This vertical comprises two complementary aspects: \emph{citation faithfulness}, which evaluates whether claims are accurately supported by their cited sources (intrinsic quality), and \emph{factual correctness}, which evaluates whether claims are true with respect to external world knowledge regardless of citation support. 
%Both are necessary to establish trust in deep research outputs.

\subsection{Diagnosing the Evaluation Landscape}\label{subsec:existing_benchmarks}

\begin{figure*}[t]
    \centering
    \includegraphics[trim=4.8cm 4.54cm 4.5cm 4.54cm,width=\linewidth]{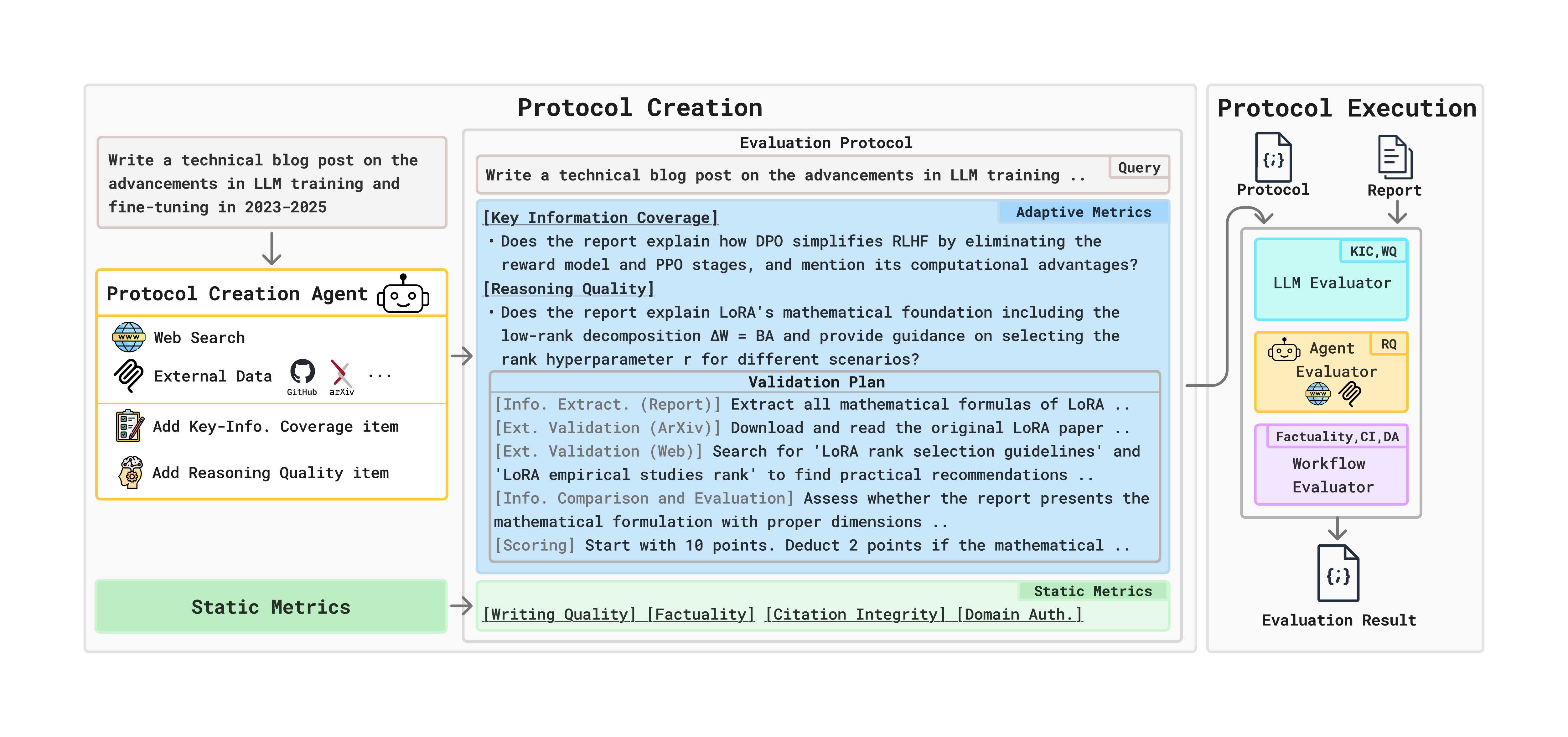}
    \caption{\textbf{\AlgoName{} Overview.} Our framework operates in two phases. \textbf{Left:} Protocol Creation, where query-independent Static Metrics are combined with Adaptive Metrics constructed by an agent equipped with web search tools and optional tools to access external data. \textbf{Right:} Protocol Execution, where each metric is routed to the appropriate evaluator, either an LLM, agent with tool access, or workflow.}
    \label{fig:dream_overview}
\vspace{-2ex}
\end{figure*}

% We map existing DRE benchmarks onto this taxonomy, examining how each vertical is evaluated and under which paradigm: human-curated annotation, closed-loop LLM-based evaluation, or citation-centered workflow-based verification. Table~\ref{tab:taxonomy} summarizes this mapping.
In \Cref{tab:taxonomy}, we map existing DRE benchmarks onto our taxonomy and group them by evaluation paradigm—human-curated, closed-loop LLM-based, or citation-centered workflow-based—based on how their metrics are \emph{created} and \emph{executed}.

% \paragraph{Human-Curated and Dataset-Specific Benchmarks.}
% Human-curated benchmarks emphasize reliability through carefully designed rubrics and annotations. ResearchRubrics~\cite{sharma2025researchrubrics} evaluates all four verticals using manually constructed criteria, requiring over 2,800 hours of human effort. DeepResearchGym~\cite{coelho2025deepresearchgym} evaluates task compliance through dataset-specific proxies, such as whether reports cover documents historically clicked by users. LiveResearchBench~\cite{wang2025liveresearchbench} combines LLM-generated checklists with human-in-the-loop verification. While these approaches achieve high annotation quality, they are inherently expensive, non-scalable, and tightly coupled to specific datasets or domains.

\vspace{-0.4em}
\paragraph{Human-Defined Evaluation Criteria.}
This paradigm includes benchmarks with human-defined evaluation criteria, specified either through manual rubrics or dataset-specific proxies. While reliable and interpretable, these approaches tightly couple evaluation to specific datasets, domains, or costly curation.
ResearchRubrics~\cite{sharma2025researchrubrics} evaluates all four verticals using manually constructed rubrics, requiring over 2,800 hours of expert annotation. DeepResearchGym~\cite{coelho2025deepresearchgym} encodes task compliance via dataset-specific behavioral proxies, such as whether reports cover documents historically clicked by users. LiveResearchBench~\cite{wang2025liveresearchbench} adopts a hybrid approach, combining LLM-generated checklists with human-in-the-loop verification.

% \paragraph{Closed-Loop LLM-Based Evaluation.}
% To improve scalability, several benchmarks replace human annotation with LLM-generated evaluation criteria. DeepResearch Bench~\cite{du2025deepresearch} introduces RACE, which uses an LLM to generate weighted rubrics for presentation quality, instruction following, and depth, scoring reports against a reference. DeepResearch Arena~\cite{wan2025deepresearch} removes the reference requirement via ACE, prompting an LLM to generate a query-specific checklist. However, these approaches rely on static LLMs during rubric or checklist creation. As a result, they lack access to external tools, temporal context, or independent evidence gathering, limiting their ability to evaluate analytical depth and task compliance in a grounded, up-to-date manner.
\vspace{-0.4em}
\paragraph{Closed-Loop LLM-Based Evaluation.}
To improve scalability, several benchmarks replace human annotation with LLM-generated evaluation criteria. DeepResearch Bench~\cite{du2025deepresearch} uses RACE to generate weighted rubrics for presentation quality, instruction following, and depth, while DeepResearch Arena~\cite{wan2025deepresearch} removes references via ACE, prompting an LLM to generate a query-specific checklist. However, rubric and checklist construction is performed by static LLMs, which lack access to external tools, temporal context, or independent evidence gathering. This constrains their ability to evaluate grounded reasoning and evolving task requirements.

% \paragraph{The Citation-Alignment Fallacy.}
% Workflow-based evaluation is most prevalent in the Source Quality vertical. Benchmarks such as DeepResearch Bench (FACT), DeepResearchGym (Retrieval Faithfulness), and LiveResearchBench (Citation Accuracy) employ similar pipelines: an LLM extracts claims and associated URLs, cited content is retrieved, and a secondary LLM verifies whether claims align with the retrieved sources. While effective at detecting misrepresentation of sources, these workflows primarily measure \emph{intrinsic} citation alignment rather than \emph{extrinsic} factual correctness. Because verification is restricted to provided citations, they are inherently insensitive to temporal drift and to claims supported by outdated or unreliable sources. As a consequence, a report can achieve perfect citation faithfulness—every claim accurately reflecting its cited source—while remaining factually incorrect or obsolete with respect to the external world, a concrete instantiation of the \emph{Mirage of Synthesis}.
\vspace{-0.4em}
\paragraph{Citation-Alignment Workflows.} Prevalent in the Source Quality vertical, this paradigm employs multi-step pipelines to verify alignment between claims and cited URLs. Benchmarks such as DeepResearch Bench (FACT), DeepResearchGym (Retrieval Faithfulness), and LiveResearchBench (Citation Accuracy) extract $\langle \text{claim, URL} \rangle$ pairs, retrieve the cited content, and utilize an LLM to judge alignment. While effective at detecting source misrepresentation, these workflows primarily measure \emph{intrinsic} citation faithfulness rather than \emph{extrinsic} factual correctness. Because verification is restricted to provided citations, they remain insensitive to claims supported by outdated or unreliable sources. This creates the \emph{Citation-Alignment Fallacy}, where a report can contain entirely accurate citations while still being factually incorrect or obsolete with respect to the external world.

% \paragraph{Systematic Imbalance and Evaluator Capability Mismatch.}
% Applying the taxonomy across evaluation paradigms reveals a consistent imbalance. Presentation Quality and Task Compliance are extensively evaluated across benchmarks, while Source Quality is predominantly assessed through intrinsic citation alignment. In contrast, extrinsic factual correctness, temporal validity, and grounded reasoning receive little direct evaluation. This imbalance gives rise to the \emph{Mirage of Synthesis}: reports that are fluent, well-structured, and well-cited can score highly despite being outdated, factually incorrect, or logically flawed. Crucially, this failure mode stems from a shared structural limitation rather than isolated design choices. Across paradigms, evaluators lack the same class of capabilities as the agents they assess—namely, the ability to independently retrieve external evidence, reason over competing sources, and incorporate temporal context. This evaluator capability mismatch motivates the agentic evaluation framework introduced in the following section.
\vspace{-0.4em}
\paragraph{Systematic Imbalance and Evaluator Capability Mismatch.}
Applying the taxonomy reveals a consistent imbalance across evaluation paradigms. Presentation Quality and Task Compliance are extensively evaluated, while Source Quality is predominantly assessed through intrinsic citation alignment. In contrast, extrinsic factual correctness, temporal validity, and grounded reasoning receive little direct evaluation. This imbalance gives rise to the \emph{Mirage of Synthesis}, where surface-level fluency and citation alignment are evaluated, while factual, temporal, and logical defects remain systematically unassessed. Crucially, this failure stems from a structural \textbf{capability mismatch}, where across all paradigms, evaluators lack critical capabilities available to the agents they assess, namely, to independently retrieve evidence, reason over competing sources, and incorporate temporal context.
% where across all paradigms, evaluators lack critical capabilities available to the agents they assess, namely, the ability to independently retrieve evidence, reason over competing sources, and incorporate temporal context.

\section{\AlgoName: DRE with Agentic Metrics}\label{sec:dream}

% The analysis in \Cref{sec:taxonomy} reveals that current benchmarks systematically fail when evaluation criteria are static, citation-bound, or temporally unaware. These failures stem from an \emph{evaluator capability mismatch}: static observers cannot verify what they cannot independently research. To address this, we introduce \AlgoName{}, a framework that enforces \textbf{capability parity} by making the evaluation process itself agentic. Given a research query, \AlgoName{} constructs a query-specific evaluation \emph{protocol} (\cref{sec:recipe_creation}) and executes it using specialized evaluators (\cref{sec:recipe_execution}), as illustrated in \Cref{fig:dream_overview}.

To address the capability mismatch identified in \Cref{sec:taxonomy}, we introduce \AlgoName{}, a framework that enforces \textbf{capability parity} by making the evaluation process itself agentic. Given a research query, \AlgoName{} constructs a query-specific \emph{evaluation protocol} (\Cref{sec:recipe_creation}) and executes it using specialized evaluators (\Cref{sec:recipe_execution}), as illustrated in \Cref{fig:dream_overview}.

\subsection{Phase 1: Protocol Creation}\label{sec:recipe_creation}
% Because deep research questions admit multiple valid report realizations, evaluation criteria cannot rely on a single reference or fixed rubric. \AlgoName{} addresses this by constructing a query-specific \emph{evaluation protocol} that defines task-relevant evaluation criteria independently of any particular report. The protocol balances cross-task consistency via static metrics with adaptive, evidence-grounded metrics derived from external sources.
Because deep research questions admit multiple valid realizations, evaluation cannot rely on a single reference or a fixed rubric. \AlgoName{} addresses this by constructing a query-specific \emph{evaluation protocol} that defines task-relevant evaluation criteria independently of any particular report. The protocol consists of two complementary classes of metrics: Static Metrics for universal quality standards, and Adaptive Metrics for task-specific depth. Full implementation details for the metric suite are provided in Appendix~\ref{app:metrics_details}.

\begin{figure*}[t]
    \includegraphics[clip, trim=4.55cm 4.54cm 4.54cm 4.56cm,width=\linewidth]{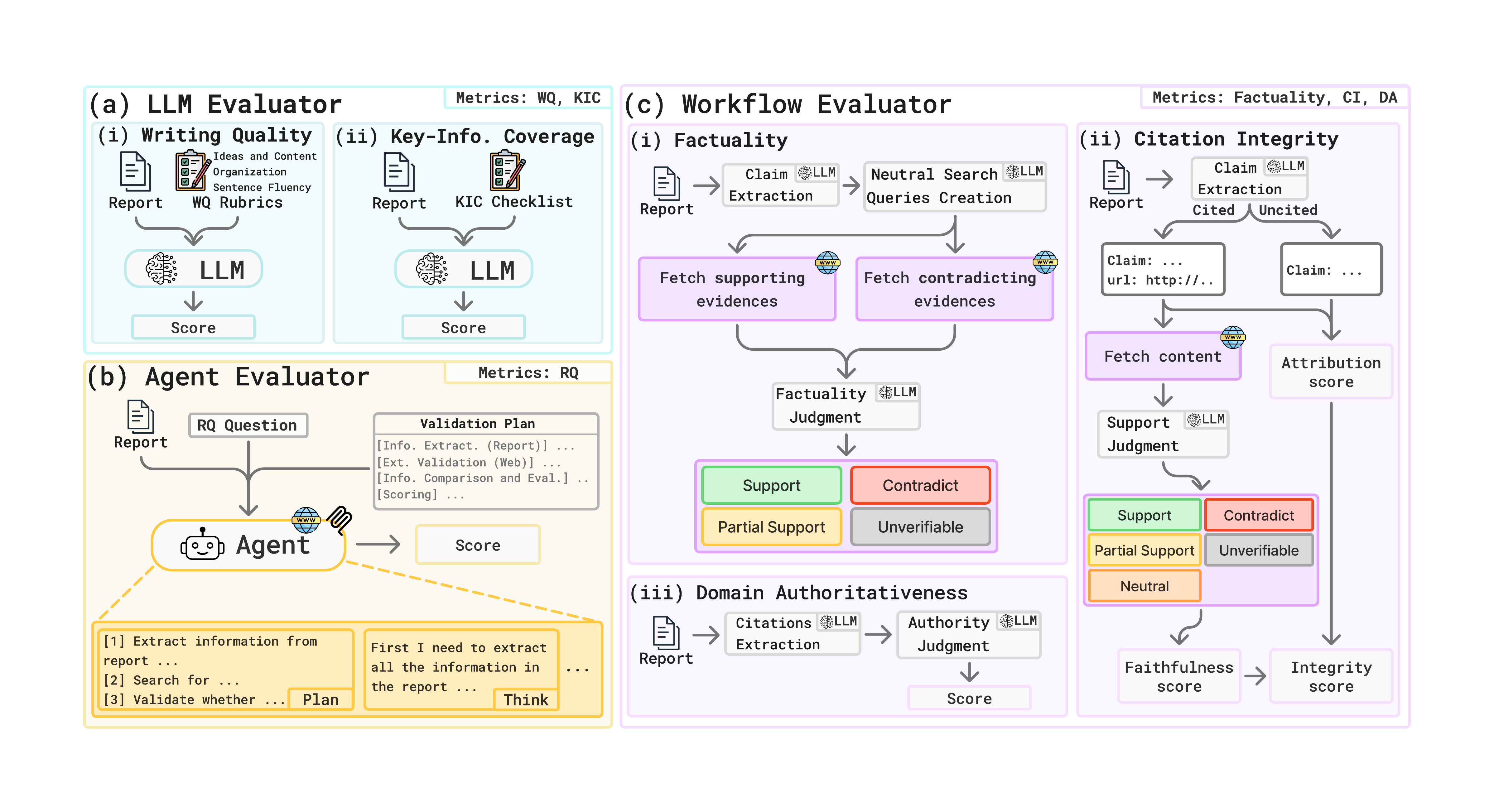}
    \caption{\textbf{\AlgoName{} Protocol Execution Evaluators. } (a) LLM Evaluator assesses writing quality (WQ) and key-information coverage (KIC); (b) Agent Evaluator evaluates reasoning quality (RQ) using external tools; (c) Workflow Evaluator performs factuality assessment via evidence retrieval, citation integrity (CI) verification through claim-source validation, and domain authoritativeness (DA) scoring via credibility assessment of extracted citations.} 
    \label{fig:method_evaluators}
\vspace{-3ex}
\end{figure*}

\paragraph{Static Metrics.} These query-agnostic criteria apply uniformly across tasks: (i) \emph{Writing Quality (WQ)} assesses presentation quality along the fixed dimensions of Ideas and Content, Organization, and Sentence Fluency; (ii) \emph{Factuality} validates claims against external knowledge independent of citation support; (iii) \emph{Citation Integrity (CI)} verifies that claims are attributed to sources and supported by the cited content; and (iv) \emph{Domain Authoritativeness (DA)} evaluates cited sources credibility to ensure reliance on reputable, high-quality domains.

\paragraph{Adaptive Metrics.} To capture query-dependent expectations, we instantiate a \emph{Protocol Creation Agent} as a CodeAgent \citep{wang2024executable} equipped with retrieval tools including web search, ArXiv, and GitHub. 
% Given a query, the agent performs a lightweight tool selection step to reduce noise, then conducts research to construct two metrics:
Given a query, the agent performs lightweight tool selection to reduce noise, then conducts research to construct two metrics:
\begin{itemize}[leftmargin=3.3mm, labelsep=0.5em] 
    \item \textbf{Key-Information Coverage (KIC):} The agent identifies essential facts by retrieving up-to-date sources and converting each key point into a verifiable \emph{yes/no question}. This transforms coverage assessment into a grounded, temporally aware checklist that flags missing or outdated content.
    
    \item \textbf{Reasoning Quality (RQ):} The agent generates query-specific questions paired with \emph{structured validation plans}. These plans specify the information to extract from both the report and independent sources, detailing how to cross-reference findings to ensure analytical depth is evaluated based on substantive reasoning. RQ is illustrated in the left panel of \Cref{fig:teaser} and in \Cref{fig:dream_overview}.
\end{itemize}

% \subsection{Mapping Taxonomy Gaps to DREAM Metrics}

% The design of \AlgoName{} directly operationalizes the taxonomy introduced in \Cref{sec:taxonomy}. Each metric targets a specific evaluation gap exposed by existing benchmarks and is paired with an evaluator whose capabilities are sufficient for the task. The resulting taxonomy-to-metric mapping is as follows:

% \begin{itemize}[leftmargin=1.2em, itemsep=3pt, topsep=4pt]
%   \item \textbf{Presentation Quality} (surface fluency dominates): Writing Quality (WQ), evaluated by an LLM.
%   \item \textbf{Task Compliance} (static criteria miss query needs): Key-Information Coverage (KIC), evaluated via an adaptive LLM checklist.
%   \item \textbf{Analytical Depth} (ungrounded reasoning): Reasoning Quality (RQ), evaluated by a tool-using agent.
%   \item \textbf{Source Quality (Intrinsic)} (citation misrepresentation): Citation Faithfulness (CF), evaluated via a claim--source verification workflow.
%   \item \textbf{Source Quality (Extrinsic)} (citation-aligned falsehoods; temporal drift): Factuality, evaluated via retrieval-grounded verification.
% \end{itemize}

\subsection{Phase 2: Protocol Execution}\label{sec:recipe_execution}

Once a protocol is constructed, \AlgoName{} executes each metric using an evaluator whose capabilities match the metric’s requirements. Evaluator selection follows the capability parity principle, i.e., each metric is routed to the simplest evaluator that possesses the capabilities required to close the diagnosed gap. As shown in \Cref{fig:method_evaluators}, \AlgoName{} employs three evaluator types.

% \begin{figure}[t]
%     \includegraphics[width=\linewidth]{figures/method_evaluators.pdf}
%     \caption{\textbf{\AlgoName{} Protocol Execution}. (a) LLM Evaluator assesses writing quality (WQ) and key-information coverage (KIC); (b) Agent Evaluator evaluates reasoning quality (RQ) using external tools; (c) Workflow Evaluator performs factuality assessment via evidence retrieval and citation faithfulness (CF) verification through claim-source validation.} 
%     \label{fig:method_evaluators}
% \vspace{-3ex}
% \end{figure}

\paragraph{LLM Evaluator.}
The LLM Evaluator is used for metrics that require judgment but not external tool use. It assesses Writing Quality using fixed rubrics to ensure calibration and evaluates Key-Information Coverage by verifying report content against the agent-generated yes/no checklist.

\paragraph{Agent Evaluator.}
The Agent Evaluator, instantiated as a CodeAgent, executes the Reasoning Quality metric. Unlike KIC, which focuses on coverage, RQ probes the coherence and validity of reasoning across the report. The agent autonomously follows the validation plan created in phase 1, retrieving external evidence as needed and assigning a final score based on evidentiary support.

\paragraph{Workflow Evaluator.}
% The Workflow Evaluator implements two complementary verification pipelines. For \emph{Citation Faithfulness}, it extracts \{claim, source\} pairs and verifies alignment.
% %, utilizing a memory pool to handle cumulative context in long-form reports.
% For \emph{Factuality}, claims are evaluated independently of citations by generating \textbf{neutralized search queries} to retrieve external evidence from the web. This enables verification of both uncited claims and cited claims whose sources are outdated, unreliable, or inconsistent with external ground-truth knowledge. Together, these pipelines ensure that CF detects source misrepresentation, while Factuality captures extrinsic factual errors that citation-based methods inherently miss. 

% The Workflow Evaluator implements two complementary verification pipelines. For \emph{Citation Faithfulness}, it extracts claim-source pairs and verifies alignment. For \emph{Factuality}, claims are evaluated independently of citations by generating \textbf{neutralized search queries} to retrieve external evidence. This enables verification of uncited claims and cited claims whose sources are outdated, unreliable, or inconsistent with external knowledge. Together, these pipelines ensure that CF detects source misrepresentation, while Factuality captures extrinsic factual errors that citation-based methods inherently miss.
The Workflow Evaluator implements three complementary verification pipelines. For Citation Integrity, the system computes the harmonic mean of \emph{Claim Attribution} (the ratio of cited to total verifiable claims) and \emph{Citation Faithfulness} (the alignment of cited claims with their source content). For Factuality, claims are extracted and evaluated independently of citations by generating \textbf{neutralized search queries} to retrieve external evidence. Finally, for Domain Authoritativeness, the extracted citations are assessed by scoring the reputation and reliability of referenced domains.
Together, this suite distinguishes between internal citation adherence (CI) and extrinsic empirical truth (Factuality), ensuring that analytical depth is grounded in credible domains (DA).

\section{Validation of \AlgoName{}}\label{sec:validation}
% We validate \AlgoName{} through a set of complementary validation studies, each designed to assess a different aspect of the evaluation framework. 
% First, we validate the quality of the agent-constructed evaluation protocols themselves using human judgment, establishing that the generated criteria are interpretable, relevant, and verifiable (\Cref{sec:human_eval}). 
% Next, we conduct targeted sensitivity analyses that isolate specific failure modes identified by our taxonomy, including temporal degradation, reasoning flaws, and extrinsic factual errors (\Cref{sec:temporal,sec:reasoning,sec:factuality}). 
% Finally, we evaluate whether writing quality can be assessed in a reference-free manner while remaining aligned with human judgment (\Cref{sec:writing}).
% Unless otherwise specified, all experiments use queries from DeepResearch and Claude-Sonnet-4.5~\citep{anthropic2025claudesonnet} as the base LLM judge.

We evaluate \AlgoName{} through a set of complementary validation studies targeting distinct aspects of the evaluation framework. First, we assess the agent-constructed protocols via human judgment, establishing that the generated criteria are interpretable, relevant, and verifiable (\Cref{sec:human_eval}). Next, we conduct targeted sensitivity analyses that isolate specific failure modes identified by our taxonomy, including temporal degradation, reasoning flaws, and extrinsic factual errors (\Cref{sec:temporal,sec:reasoning,sec:factuality}). 
Finally, we demonstrate that writing quality can be assessed in a reference-free manner while remaining aligned with human judgment (\Cref{sec:writing}). 
Unless otherwise specified, we use queries from DeepResearch Bench (DRB) and Claude Sonnet 4.5~\citep{anthropic2025claudesonnet} as the base LLM.

\subsection{Human Evaluation of Protocol Quality}\label{sec:human_eval}
% A central claim of \AlgoName{} is that adaptive evaluation criteria can be constructed agentically, replacing manually authored rubrics. We therefore evaluate the quality of the agent-generated protocol components—\emph{Key-Information Coverage (KIC)} and \emph{Reasoning Quality (RQ)}—and ablate the protocol construction process to isolate the contribution of agentic structuring and external retrieval.
% Specifically, this study assesses whether the generated criteria align with human judgments of relevance, clarity, and verifiability, and whether agentic design choices are necessary to achieve this alignment.

% We conducted a human study where expert and non-expert annotators rated KIC items on relevance, clarity, and verifiability, and RQ items on the same dimensions as well as validation plan validity. Ratings used a 1--3 scale, normalized to $[0,1]$ for aggregation. Full details are provided in Appendix~\ref{app:human_eval}.

% As shown in Table~\ref{tab:protocol-human}, annotators rated the protocol criteria generated by the full agent highly (KIC: 0.92, RQ: 0.93), with RQ clarity (0.97) and plan validity (0.99) scoring particularly well. A three-way ablation reveals that agentic structuring alone improves over a plain LLM (RQ average: 0.84 vs.\ 0.70). Furthermore, adding retrieval significantly boosts performance, most notably increasing verifiability (KIC: 0.75 $\rightarrow$ 0.91, RQ: 0.67 $\rightarrow$ 0.80) and RQ plan validity (0.92 $\rightarrow$ 0.99). These results confirm the value of both structured multi-step reasoning and grounding in external evidence.

A central claim of \AlgoName{} is that adaptive evaluation criteria can be constructed agentically, replacing manually authored rubrics. We therefore evaluate the quality of the agent-generated protocol metrics, KIC and RQ. 
We conducted a human study where expert and non-expert annotators rated generated items on relevance, clarity, and verifiability, as well as plan validity for RQ, using a 1--3 scale normalized to $[0,1]$ (details are provided in Appendix~\ref{app:human_eval}). We further performed a three-way ablation to isolate the impact of agentic structuring and retrieval.

As shown in Table~\ref{tab:protocol-human}, annotators rated the protocol criteria generated by the full agent highly (KIC: 0.92, RQ: 0.93), with RQ clarity (0.97) and plan validity (0.99) scoring particularly well. The ablation reveals that agentic structuring alone improves over a plain LLM (RQ average: 0.84 vs.\ 0.70). Furthermore, adding retrieval significantly boosts performance, notably increasing verifiability (KIC: 0.75 $\rightarrow$ 0.91, RQ: 0.67 $\rightarrow$ 0.80) and RQ plan validity (0.92 $\rightarrow$ 0.99). These results confirm the value of both structured multi-step reasoning and grounding in external evidence.

\begin{table}[t]
\centering
\caption{\textbf{Human evaluation of KIC and RQ items.} Performance by metric and method. The agent with retrieval achieves the best performance across all axes.}
\resizebox{\columnwidth}{!}{%
\begin{tabular}{@{}l l ccccc@{}}
\toprule
\textbf{Metric} & \textbf{Method} & \textbf{Rel.} & \textbf{Verif.} & \textbf{Clar.} & \textbf{Valid.} & \textbf{Avg.} \\
\midrule
\multirow{3}{*}{\textbf{KIC}} 
 & LLM     & 0.84 & 0.73 & 0.80 & -- & 0.79 \\
 & Agent         & 0.88 & 0.75 & 0.87 & -- & 0.83 \\
 & Agent + Ret.  & \textbf{0.94} & \textbf{0.91} & \textbf{0.92} & -- & \textbf{0.92} \\
\midrule
\multirow{3}{*}{\textbf{RQ}} 
 & LLM     & 0.82 & 0.51 & 0.78 & 0.70 & 0.70 \\
 & Agent         & 0.89 & 0.67 & 0.88 & 0.92 & 0.84 \\
 & Agent + Ret.  & \textbf{0.94} & \textbf{0.80} & \textbf{0.97} & \textbf{0.99} & \textbf{0.93} \\
\bottomrule
\end{tabular}%
}
\label{tab:protocol-human}
\vspace{-2ex}
\end{table}

\subsection{Temporal Awareness in KIC}\label{sec:temporal}
We next evaluate the framework's sensitivity to \emph{temporal obsolescence}, a core failure mode identified by our taxonomy. When temporal validity is central to a query, a competent evaluator should consistently penalize reports generated with outdated knowledge. We investigated this by selecting 20 temporally volatile queries (e.g., ``TikTok US legal status'') and generating three report variants for each: one using current information (Dec 2025), and two simulated with knowledge cutoffs of Jan 2025 and Jan 2024. We then evaluated all reports using both \AlgoName{}--KIC and DRB--RACE (Comprehensiveness and Insight), as reported in Table~\ref{tab:temporal_sensitivity}.

\begin{figure*}[t]
    \includegraphics[width=\linewidth]{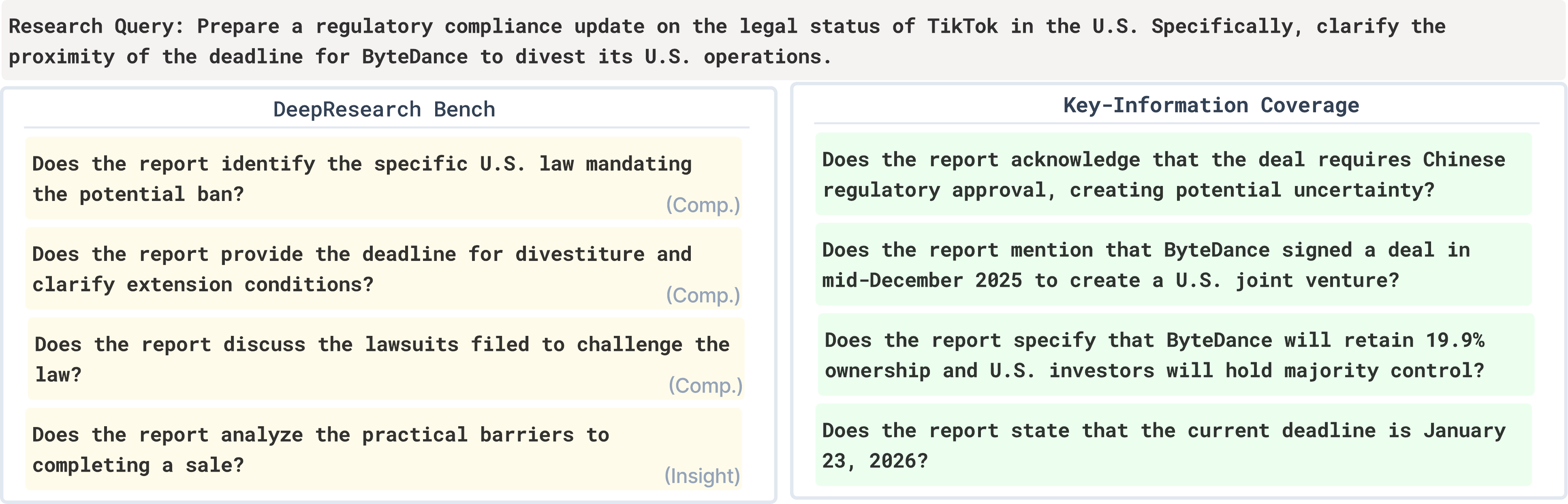}
    \caption{\textbf{Temporal Awareness in KIC Evaluation.} Comparison of evaluation criteria for a TikTok legal status query, showing DeepResearch Bench's static criteria (left) versus DREAM's KIC criteria (right) that incorporate time-sensitive facts (e.g., mid-December 2025 joint venture deal and January 23, 2026 deadline).}
    \label{fig:timeaware_example}
\vspace{-2ex}
\end{figure*}

The results reveal a sharp contrast. DRB--RACE exhibits weak temporal sensitivity: Comprehensiveness fails to penalize moderately outdated reports (50.02 $\rightarrow$ 50.04 for Jan 2025), while Insight degrades only marginally (50.54 $\rightarrow$ 48.78). In contrast, \AlgoName{}--KIC degrades monotonically with information staleness, dropping from 79.35 (current) to 44.80 (Jan 2025) and 22.34 (Jan 2024). This performance gap is structural; DRB--RACE evaluates reports against static criteria (e.g., identifying a relevant law), which remain satisfiable even with obsolete facts. KIC instead encodes time-sensitive expectations derived from up-to-date evidence (\Cref{fig:timeaware_example}), causing outdated reports to fail explicit verification. Thus, effective temporal awareness requires agentic evaluation rather than static rubrics.

% \subsection{Detecting Reasoning Flaws}\label{sec:reasoning}

% This experiment evaluates whether an evaluation framework can distinguish \emph{reasoning soundness} from surface-level plausibility, a key failure mode under the Analytical Depth vertical of our taxonomy. Static evaluators that rely on holistic judgments or stylistic cues often fail to penalize flawed reasoning when reports remain fluent and well-structured. We test whether agentic reasoning validation can reliably surface analytical failures that static, reference-based evaluators overlook.

% To validate RQ's ability to detect analytical failures, we conducted a controlled experiment using ten research queries spanning domains where reasoning quality is critical (e.g., policy analysis, technical comparisons, causal explanations). For each query, we generated two reports: a standard version and a malformed variant containing deliberately injected flaws, such as unsupported causal claims or circular reasoning, while preserving surface-level plausibility and fluency. Full experimental details are provided in Appendix~\ref{subapp:reasoning}.

% \Cref{fig:rqe_vs_race} plots relative score degradations between well-reasoned and malformed reports. DRB--RACE produces only $\approx 10\%$ average degradation, with a mode below zero---meaning malformed reports sometimes score higher. In contrast, DREAM--RQ exhibits a clear signal centered around $\approx 40\%$ degradation, reliably surfacing reasoning failures that are otherwise masked by the report's stylistic coherence.

\subsection{Detecting Reasoning Flaws}\label{sec:reasoning}
% In deep research reports, analytical failures often occur within fluent, well-structured, and superficially coherent narratives. As a result, detecting flawed reasoning requires evaluation methods that go beyond surface plausibility and explicitly probe the validity of reasoning chains. This subsection examines whether existing evaluation approaches are capable of distinguishing well-reasoned reports from those containing subtle but substantive reasoning errors under such conditions.

Analytical failures in deep research reports often occur within fluent, well-structured, and superficially coherent narratives. Detecting such flawed arguments requires evaluation methods that look beyond surface plausibility to explicitly probe reasoning validity. We examined whether existing metrics can distinguish sound reports from those containing subtle but substantive reasoning errors.

% To study this, we design a controlled experiment that isolates reasoning quality while holding surface fluency constant. We select ten research queries spanning domains where analytical rigor is essential (\textit{e.g.}, policy analysis, technical comparisons, and causal explanations). For each query, we generate two reports: a standard version and a malformed variant containing deliberately injected reasoning flaws, such as unsupported causal claims or circular arguments, while preserving overall structure and linguistic quality. Full experimental details are provided in Appendix~\ref{subapp:reasoning}.

To study this, we designed a controlled experiment that isolates reasoning quality while holding surface fluency constant. We selected 10 complex queries spanning domains where analytical rigor is essential (e.g., policy analysis and technical comparisons), and generated two report variants for each: a standard version and a malformed variant. The malformed variant contained deliberately injected reasoning flaws, such as circular arguments or unsupported claims, while preserving a fluent structure (see full details in Appendix~\ref{subapp:reasoning}).

% \Cref{fig:rqe_vs_race} reports the relative score degradation assigned to malformed reports. DRB--RACE exhibits weak and inconsistent sensitivity, with an average degradation of approximately $10\%$ and a mode below zero, indicating that malformed reports can sometimes receive higher scores than well-reasoned ones. In contrast, DREAM--RQ produces a consistent signal, centered around a $40\%$ degradation, reliably penalizing reports with flawed reasoning despite their surface coherence.

\begin{figure}[t]
    \centering
    \includegraphics[clip, trim=4.6cm 4.54cm 4.54cm 4.56cm,width=\linewidth]{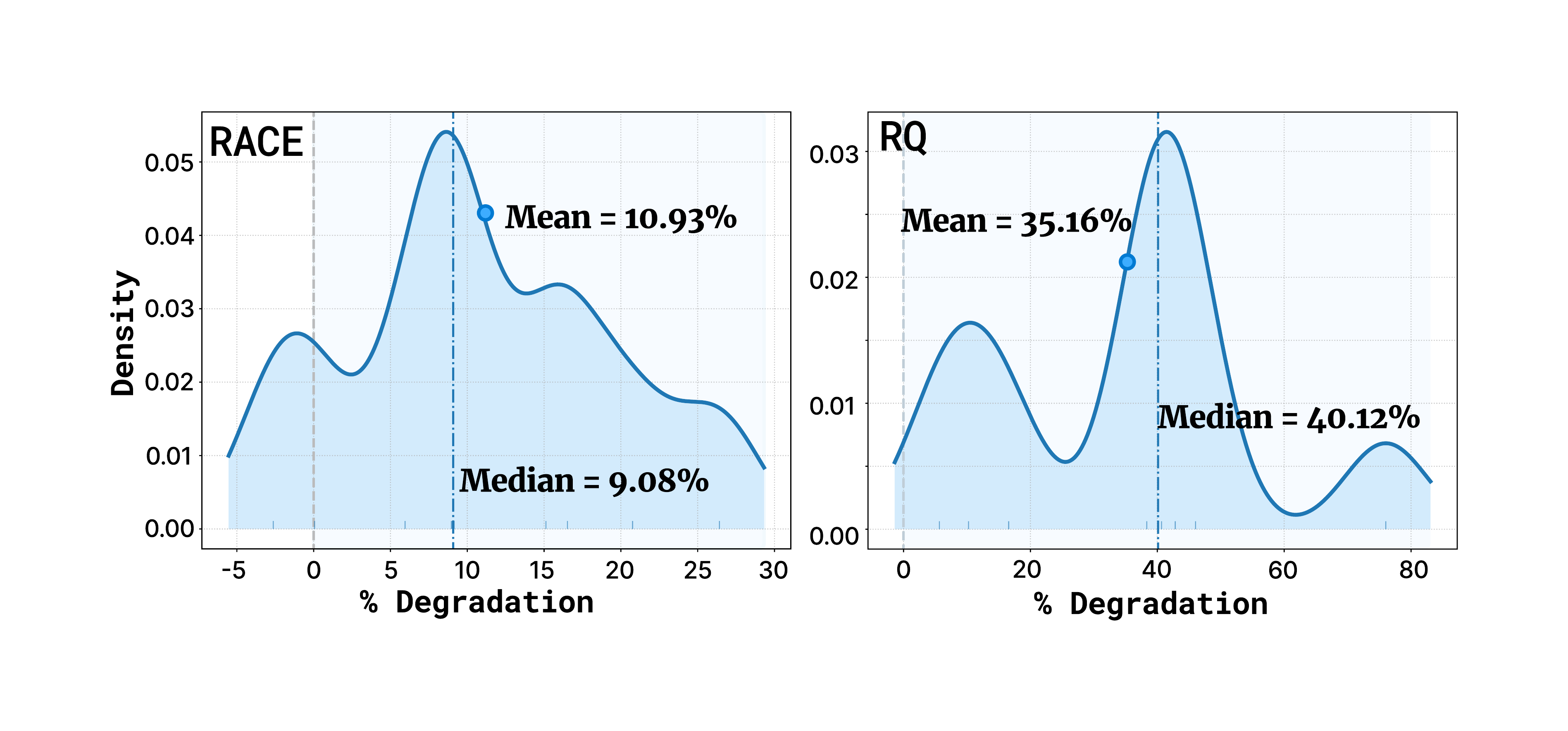}
    \caption{\textbf{Reasoning flaws detection.} Relative score degradations between well-reasoned and malformed reports. DREAM--RQ centers around $40.1\%$ degradation, while RACE centers around $9.1\%$, with several malformed reports outscoring well-reasoned ones.}
    \label{fig:rqe_vs_race}
    \vspace{-2ex}
\end{figure}

\Cref{fig:rqe_vs_race} reports the relative score degradation of malformed reports, revealing a significant gap. DRB--RACE exhibits weak and inconsistent sensitivity, with an average degradation of $\sim\!9\%$, and frequently scores malformed reports higher than well-reasoned ones. By comparison, \AlgoName{}--RQ produces a consistent signal, centered at $\sim\!40\%$ degradation, reliably penalizing reports with flawed reasoning despite their surface coherence.

% \subsection{Grounding Beyond Citation Alignment}\label{sec:factuality}

% This experiment evaluates whether an evaluation framework can detect \emph{extrinsic factual errors}: claims that are incorrect with respect to the external world, even if they appear citation-aligned. Citation faithfulness metrics verify internal consistency but are blind to false or outdated information supported by unreliable sources. We test whether retrieval-grounded factual verification resolves this.

% To validate the proposed Factuality metric, we conducted a controlled injection study using 15 pairs of claims detailed in Appendix~\ref{subapp:factuality_beyond_CF}, each contrasting a factually correct statement (e.g., ``Humans use virtually 100\% of their brains") with a plausible but incorrect variant (e.g., ``Most humans only use 10\% of their brain capacity"). We swept the corruption rate $r$ from 0 to 1, progressively replacing correct claims with their incorrect counterparts, each supported by a matching URL. As shown in \Cref{fig:teaser} (middle), DRB--FACT scores remain flat across all corruption levels. Since the metric validates claims solely against the provided (incorrect) reference text, it is vulnerable to the \emph{Mirage of Synthesis}. In contrast, DREAM--Factuality degrades monotonically, closely tracking the true corruption rate. This demonstrates that only metrics verifying claims against external knowledge can reliably detect plausible hallucinations.

\subsection{Grounding Beyond Citation Alignment}\label{sec:factuality}
Claims without citations are inherently invisible to citation-alignment workflows, motivating a reference-free factual verification metric such as \AlgoName{}--Factuality. More critically, cited claims can be faithful to their sources while remaining factually incorrect with respect to the external world.

We isolate this failure mode with a controlled corruption study. Starting from a base set of factual claims, we construct pairs consisting of a correct version with a supporting citation, and a plausible but incorrect variant with a matching citation (15 pairs; cf. \Cref{subapp:factuality_beyond_CF}). We then sweep a corruption rate $r\!\in\![0,1]$, replacing an $r$ fraction of correct claims with incorrect ones while preserving citation alignment by construction.

As shown in \Cref{fig:teaser} (middle), DRB--FACT scores remain invariant across all corruption levels, as the metric validates claims solely against the provided sources. In contrast, \AlgoName{}--Factuality degrades monotonically with increasing $r$, closely tracking the true error rate. This demonstrates that citation alignment is insufficient for factual evaluation, as detecting plausible, well-cited falsehoods requires access to external world knowledge.

% \begin{table}[t]
% \centering
% \caption{\textbf{Sensitivity to factual errors.} Scores normalized to clean baseline as corruption ratio increases. DRB--FACT remains insensitive; DREAM--Factuality tracks the injected error rate.}
% \begin{tabular}{c|cc}
% \toprule
% \makecell{\textbf{Corruption} \\ \textbf{Ratio}} & \makecell{\textbf{DREAM} \\ \textbf{Factuality}} & \makecell{\textbf{DRB} \\ \textbf{FACT}} \\
% \midrule
% 0.0 & 1.000 & 1.000 \\
% 0.2 & 0.757 & 0.980 \\
% 0.4 & 0.459 & 1.048 \\
% 0.6 & 0.280 & 0.895 \\
% 0.8 & 0.152 & 1.001 \\
% 1.0 & 0.064 & 0.963 \\
% \bottomrule
% \end{tabular}
% \vspace{-1em}
% \label{tab:robustness_analysis}
% \end{table}

\subsection{Reference-Free Presentation Evaluation}\label{sec:writing}
% Finally, we evaluate whether Presentation Quality can be reliably assessed without a reference report. Reference-based evaluation introduces systematic bias when references are suboptimal or outdated, often penalizing valid alternative formulations. We compare \AlgoName{}'s reference-free Writing Quality scores against human-evaluated readability scores from DeepResearch Bench, using 300 reports generated by six agents across 50 queries.

% We compute the rank correlation using Kendall's $\tau$ between the rankings induced by \AlgoName{}--WQ score and the human ground truth, yielding an average $\tau=0.6$. Since human inter-annotator agreement for subjective readability typically falls in the $0.5$--$0.7$ range, this result confirms that \AlgoName{} provides a reliable, scalable proxy for human judgment that eliminates the need for reference reports.

Finally, we evaluate whether Presentation Quality can be reliably assessed without a reference report. Reference-based evaluation may introduce bias when references are suboptimal or outdated, often penalizing valid alternative formulations. To validate the reliability of \AlgoName{}'s reference-free WQ metric, we measure its alignment with readability rankings established by DRB. Specifically, we utilize the DRB-RACE scores provided for 300 reports generated by six agents across 50 queries, which have been previously validated as a high-fidelity proxy for human preference.

We compute the rank correlation using Kendall's $\tau$ between the rankings induced by \AlgoName{}'s WQ score and the DRB-RACE rankings. This analysis yields an average $\tau=0.6$. Given that human inter-annotator agreement for subjective readability typically falls in the $0.5$--$0.7$ range, this result confirms that \AlgoName{} provides a reliable signal comparable to established benchmarks.

\section{Benchmarking Leading DRAs}
\label{sec:benchmarking}
Having validated \AlgoName{}, we now apply it to benchmark leading DRAs. Unlike current evaluations that rely on dataset-specific annotations, our framework is dataset-agnostic, enabling a unified comparison across diverse tasks. We employ three diverse datasets. \textsc{DeepResearch Bench}~\citep{du2025deepresearch} is a bilingual benchmark comprising 50 English and 50 Chinese PhD-level questions across 22 research fields; we consider the English subset. \textsc{LiveResearchBench}~\citep{wang2025liveresearchbench} focuses on timely information synthesis requiring access to recent sources, and we utilize the 80 publicly available queries.\footnote{\url{https://huggingface.co/datasets/Salesforce/LiveResearchBench}} Finally, ResearchRubrics~\citep{sharma2025researchrubrics} provides 101 queries accompanied by expert-crafted rubrics.

We compare three open-source systems: LangChain Open Deep
Research~\cite{langchain_open_deep_research} with GPT-5 as backbone
model, Smolagents Open Deep Research~\cite{roucher2025open} with Claude
Opus 4.6 as its backbone, and Tongyi Deep
Research~\cite{team2025tongyi}. For brevity, full results are deferred
to Appendix~\ref{sec:dream_benchmarking}
(Table~\ref{tab:dra_results}).

% \begin{table*}[t]
% \centering
% \caption{\textbf{\AlgoName{} evaluation.} Aggregate scores for leading DRAs.}
% \fontsize{9.7pt}{12.5pt}\selectfont
% \begin{tabular}{l|cccc|cc|c}
% \toprule
% & \multicolumn{4}{c|}{\textbf{Static Metrics}} & \multicolumn{2}{c|}{\textbf{Adaptive Metrics}} & \\
% \textbf{DRA} & \textbf{WQ} & \textbf{Fact.} & \textbf{CI} & \textbf{DA} & \textbf{KIC} & \textbf{RQ} & \textbf{Avg.} \\
% \midrule
% Gemini-3-Pro DR & \textbf{69.06} & 55.20 & 64.35 & 71.59 & \textbf{71.70} & \textbf{65.92} & 66.30 \\
% OpenAI DR & 65.63 & \textbf{60.30} & \textbf{78.30} & \textbf{72.76} & 66.42 & 60.28 & \textbf{67.28} \\
% Perplexity DR & 66.11 & 53.18 & 70.68 & 71.01 & 61.88 & 58.94 & 63.63 \\
% \bottomrule
% \end{tabular}
% \vspace{-2ex}
% \label{tab:dra_results_aggregate}
% \end{table*}

% \vspace{-0.5ex}
Results reveal several notable patterns across open-source DRAs. All
three agents exhibit critically low Citation Integrity scores, exposing
a systemic weakness in citation grounding, though the failure modes
differ. Smolagents Open DR and Tongyi Deep Research achieve near-zero CI
(4.78 and 1.03 in aggregate, respectively) primarily because they rarely ground claims
to specific sources, reflecting negligible claim attribution rates.
LangChain Open DR attains higher CI (15.92) thanks to
substantially better attribution practices, yet is undermined by low
citation faithfulness: attributed sources frequently fail to support the
claims they are attached to (see~\Cref{fig:faithfulness_breakdown,fig:ci_scatter} for a
detailed breakdown). Beyond citation behavior, Smolagents Open DR
dominates in both source quality and synthesis, leading aggregate scores
for Factuality (58.15), Writing Quality (63.97), Key-Information
Coverage (75.95), and Reasoning Quality (69.16), achieving strong
content quality despite its near-absent citation discipline. Tongyi Deep
Research ranks second on Factuality (55.09) but trails on adaptive
metrics (aggregate RQ of 45.48), while LangChain Open DR records the
lowest Factuality (44.64) yet outperforms Tongyi on reasoning (RQ of
57.28). Overall, these results suggest that open-source DRAs can produce
informative and well-written reports, but remain fundamentally limited
by poor citation grounding, a critical gap for trustworthy research
agents.

\paragraph{Robustness across Backbone Models.} To evaluate the stability of our agentic metrics, we conducted a sensitivity analysis using different LLMs, specifically DeepSeek-V3.2~\cite{liu2025deepseek} and Kimi-K2.5~\cite{team2026kimi}, as the backbone for protocol execution on \textsc{DeepResearch Bench}. As shown in~\Cref{tab:judge_sensitivity}, our findings indicate that while absolute scores exhibit minor fluctuations depending on the judge's internal calibration—an expected variance across different LLM families—the relative performance rankings of the evaluated deep research agents remain highly robust. Notably, we observe perfect alignment across all metrics, with the slight exception of the Writing Quality (WQ) metric, for which Claude Sonnet 4.5 tended to provide more uniform scores, showing a less pronounced preference for specific agent compared to the other backbone models.

\section{Conclusion}\label{sec:conclusion}

This work introduces a unifying taxonomy for Deep Research Evaluation that identifies a fundamental failure in existing methodologies: the \emph{Mirage of Synthesis}. We demonstrate that current LLM-as-a-judge and reference-based benchmarks are often blinded by surface-level fluency and citation alignment, failing to detect deep-seated defects in factual correctness, temporal validity, and logical reasoning. These failures stem from a structural \emph{evaluator capability mismatch}, where static evaluators lack the agency required to verify the very claims they are tasked to judge.
To resolve this, we propose the principle of \textbf{capability parity} and instantiate it in \textbf{\AlgoName{}}, a framework that makes evaluation itself agentic. By transitioning from frozen rubrics to dynamic, tool-equipped evaluation protocols, \AlgoName{} provides temporally aware fact-checking and substantive reasoning validation. Our controlled evaluations confirm that these capabilities are essential for reliable assessment: \AlgoName{} detects temporal degradation and extrinsic factual errors that static benchmarks miss entirely, while reliably surfacing reasoning flaws masked by stylistic coherence. 
Ultimately, \AlgoName{} demonstrates that as AI agents gain the ability to research and reason over the open web, the frameworks used to judge them must evolve in kind. By grounding assessment in evidence and agency rather than surface form, \AlgoName{} provides a scalable and reference-free blueprint for evaluating the next generation of frontier deep research agents.

\section*{Limitations}
While \AlgoName{} addresses the \emph{evaluator capability mismatch}, several limitations remain. First, reliance on external tools introduces dependencies on third-party service availability and potential retrieval bias—a trade-off inherent to prioritizing temporal validity over closed-world consistency. Second, agentic evaluation is computationally intensive: multi-step verification and tool-interaction loops increase latency and cost compared to static judges. We view this as a necessary trade-off for scientific accuracy, though future work could explore optimization via caching or selective evaluation strategies. Finally, \AlgoName{} is a post-hoc evaluator of research outputs and does not directly assess intermediate research processes, such as search trajectory efficiency or source discovery dynamics. Extending evaluation to include process-level telemetry remains a promising future direction.

% \textcolor{gray}{This document does not cover the content requirements for ACL or any other specific venue.  Check the author instructions for information on maximum page lengths, the required ``Limitations'' section, and so on.}

% \section*{Acknowledgments}
% This document has been adapted
% by Steven Bethard, Ryan Cotterell and Rui Yan
% from the instructions for earlier ACL and NAACL proceedings, including those for
% ACL 2019 by Douwe Kiela and Ivan Vuli\'{c},
% NAACL 2019 by Stephanie Lukin and Alla Roskovskaya,
% ACL 2018 by Shay Cohen, Kevin Gimpel, and Wei Lu,
% NAACL 2018 by Margaret Mitchell and Stephanie Lukin,
% Bib\TeX{} suggestions for (NA)ACL 2017/2018 from Jason Eisner,
% ACL 2017 by Dan Gildea and Min-Yen Kan,
% NAACL 2017 by Margaret Mitchell,
% ACL 2012 by Maggie Li and Michael White,
% ACL 2010 by Jing-Shin Chang and Philipp Koehn,
% ACL 2008 by Johanna D. Moore, Simone Teufel, James Allan, and Sadaoki Furui,
% ACL 2005 by Hwee Tou Ng and Kemal Oflazer,
% ACL 2002 by Eugene Charniak and Dekang Lin,
% and earlier ACL and EACL formats written by several people, including
% John Chen, Henry S. Thompson and Donald Walker.
% Additional elements were taken from the formatting instructions of the \emph{International Joint Conference on Artificial Intelligence} and the \emph{Conference on Computer Vision and Pattern Recognition}.

% Bibliography entries for the entire Anthology, followed by custom entries
%\bibliography{anthology,custom}
% Custom bibliography entries only
\bibliography{custom}

\onecolumn
\appendix
% \section{Extended Related Work}\label{app:ext_related_work}
% While our primary focus is on the holistic evaluation of DRAs, an extensive body of work targets the isolated sub-skills that underpin this domain. For long-form text generation, benchmarks such as \textsc{LongWriter}~\citep{bai2024longwriter}, \textsc{HelloBench}~\citep{HelloBench2024}, and \textsc{WritingBench}~\citep{wu2025writingbenchcomprehensivebenchmarkgenerative} specifically evaluate models on coherence, length, and stylistic quality. For web interaction and task execution, \textsc{WebArena}~\citep{zhou2024webarenarealisticwebenvironment} and \textsc{BrowseComp}~\citep{wei2025browsecompsimplechallengingbenchmark} evaluate browsing skills and goal-directed navigation in controlled environments, while \textsc{Mind2Web 2}~\citep{gou2025mind2web} introduces ``Agent-as-a-Judge'' evaluation for long-horizon search tasks. Beyond these, benchmarks such as
% \textsc{GSM8K}~\citep{cobbe2021trainingverifierssolvemath}, 
% \textsc{MATH}~\citep{hendrycks2021measuringmathematicalproblemsolving}, \textsc{GAIA}~\citep{mialon2023gaiabenchmarkgeneralai}, \textsc{SciBench}~\citep{wang2024scibenchevaluatingcollegelevelscientific}, and  \textsc{BIG-Bench Hard}~\citep{kazemi2025bigbenchextrahard} focus on domain-specific or quantitative reasoning. While these frameworks offer valuable insights into specific agentic capabilities, they do not evaluate the holistic \emph{research} workflow that requires complex integration of autonomous retrieval, synthesis, citation, and reasoning over evolving evidence.

\section{Extended Related Work}\label{app:ext_related_work}

The rapid development of DRAs has outpaced the capabilities of existing evaluation frameworks. Our work addresses this gap by proposing a holistic evaluation protocol tailored to these complex systems.

\paragraph{Deep Research Systems.}
The domain of autonomous research has evolved from simple retrieval agents to complex systems capable of synthesizing long-form reports. In addition to the proprietary deep research systems (like those considered in \Cref{sec:benchmarking}), open frameworks have pioneered transparent architectures. \textsc{STORM}~\citep{shao2024assisting} introduced the paradigm of ``outline-driven retrieval'', using multi-perspective question asking to simulate expert dialogue. This was further extended by \textsc{Co-STORM}~\citep{jiang2024into} to include human-agent collaboration. In the open-source domain, \textsc{GPT-Researcher}~\citep{elovic2025gpt} established a baseline for parallelized plan-and-solve execution, where a static planner decomposes tasks for concurrent retrieval agents. \textsc{WebWeaver}~\citep{li2025webweaver} improved upon this with dynamic outline optimization, while \textsc{TTD-DR}~\citep{han2025deep} proposed a test-time diffusion framework for iterative report refinement. Despite this proliferation, a recent survey~\citep{huang2025deep} highlights critical limitations in current evaluation benchmarks, notably their ``restricted access to external knowledge''.

\paragraph{From Static Judges to Agentic Evaluation}
To scale evaluation beyond human annotation, recent work has coalesced around the \emph{LLM-as-a-Judge} paradigm~\citep{liu2023g,zheng2023judging}. However, these static judges typically focus on final outcomes, failing to capture the complex, multi-step decision-making inherent to autonomous agents. This limitation has spurred the evolution toward \emph{Agent-as-a-Judge} frameworks, designed to verify the execution process itself rather than just the result. \citet{zhuge2024agent} pioneered this approach for code generation, proposing recursive agentic evaluation to assess intermediate steps without excessive manual labor. More recently, \textsc{Mind2Web-2}~\citep{gou2025mind2web} extended this to the web domain, arguing that the complexity of long-horizon tasks exceeds the capacity of simple LLM calls and necessitates an agentic evaluator to trace navigation trajectories. However, these frameworks remain confined to specific domains like code generation or web navigation and have not yet been applied to the open-ended information synthesis required in deep research.

\vspace{0.5em}
\noindent
While our primary focus is on the holistic evaluation of DRAs, an extensive body of work targets the isolated sub-skills that underpin this domain, ranging from atomic factuality to long-form text generation.

\paragraph{Factuality and Atomic Verification.}
To address hallucination, recent works have focused on granular verification. \textsc{FActScore}~\citep{min2023factscore} introduced the paradigm of decomposing long-form text into atomic facts for independent verification. Similarly, \textsc{FacTool}~\citep{chern2023factool} pioneered the use of tool-augmented frameworks for detecting factual errors. However, these methods focus strictly on atomic truthfulness and do not evaluate the synthesis, argumentation, or structural coherence of a complete research report.

\vspace{-0.5ex}
\paragraph{Retrieval-Augmented Generation (RAG) and Citation.}
While standard RAG benchmarks like \textsc{RGB}~\citep{chen2024benchmarking} evaluate retrieval accuracy, more specialized frameworks focus specifically on attribution. \textsc{ALCE}~\citep{gao2023enabling} and \textsc{RAGAS}~\citep{es2024ragas} established the standards for measuring citation recall and faithfulness. However, to maintain reproducibility, these benchmarks typically rely on \textit{static snapshots} of the web or closed corpora where ground-truth documents are pre-defined. This constraint allows for recall calculation but fails to capture the complexity of open-ended deep research, which operates on the live, dynamic web. Unlike RAG tasks where the goal is often to retrieve a specific ``gold'' document, deep research admits multiple valid search trajectories. Consequently, evaluation cannot rely on retrieval recall against a fixed index, but must instead assess the extrinsic validity of autonomously discovered evidence and its synthesis into a coherent narrative. Additionally, it does not take into account verticals such as Presentation Quality or Analytical Depth.

\vspace{-0.5ex}
\paragraph{Web Interaction and Reasoning Benchmarks.}
For web interaction, \textsc{WebArena}~\citep{zhou2024webarenarealisticwebenvironment} and \textsc{BrowseComp}~\citep{wei2025browsecompsimplechallengingbenchmark} evaluate browsing skills and goal-directed navigation in controlled environments. Beyond these, benchmarks such as \textsc{GSM8K}~\citep{cobbe2021trainingverifierssolvemath}, \textsc{MATH}~\citep{hendrycks2021measuringmathematicalproblemsolving}, \textsc{GAIA}~\citep{mialon2023gaiabenchmarkgeneralai}, \textsc{SciBench}~\citep{wang2024scibenchevaluatingcollegelevelscientific}, and \textsc{BIG-Bench Hard}~\citep{kazemi2025bigbenchextrahard} focus on quantitative reasoning and domain-specific problem solving.

\vspace{-0.5ex}
\paragraph{Long-form Text Generation.}
Finally, benchmarks such as \textsc{LongWriter}~\citep{bai2024longwriter}, \textsc{HelloBench}~\citep{HelloBench2024}, \textsc{WritingBench}~\citep{wu2025writingbenchcomprehensivebenchmarkgenerative}, and \textsc{LongGenBench}~\citep{wu2025longgenbench} evaluate models specifically on coherence, length compliance, and stylistic quality. These frameworks primarily assess \textit{Presentation Quality}, often relying on static references or intrinsic LLM-based judging without external verification of the content's truthfulness.

\section{Agentic Taxonomy Pipeline}\label{app:taxonomy_pipeline}
\begin{figure*}[t]
    \centering
    \includegraphics[width=0.6\textwidth]{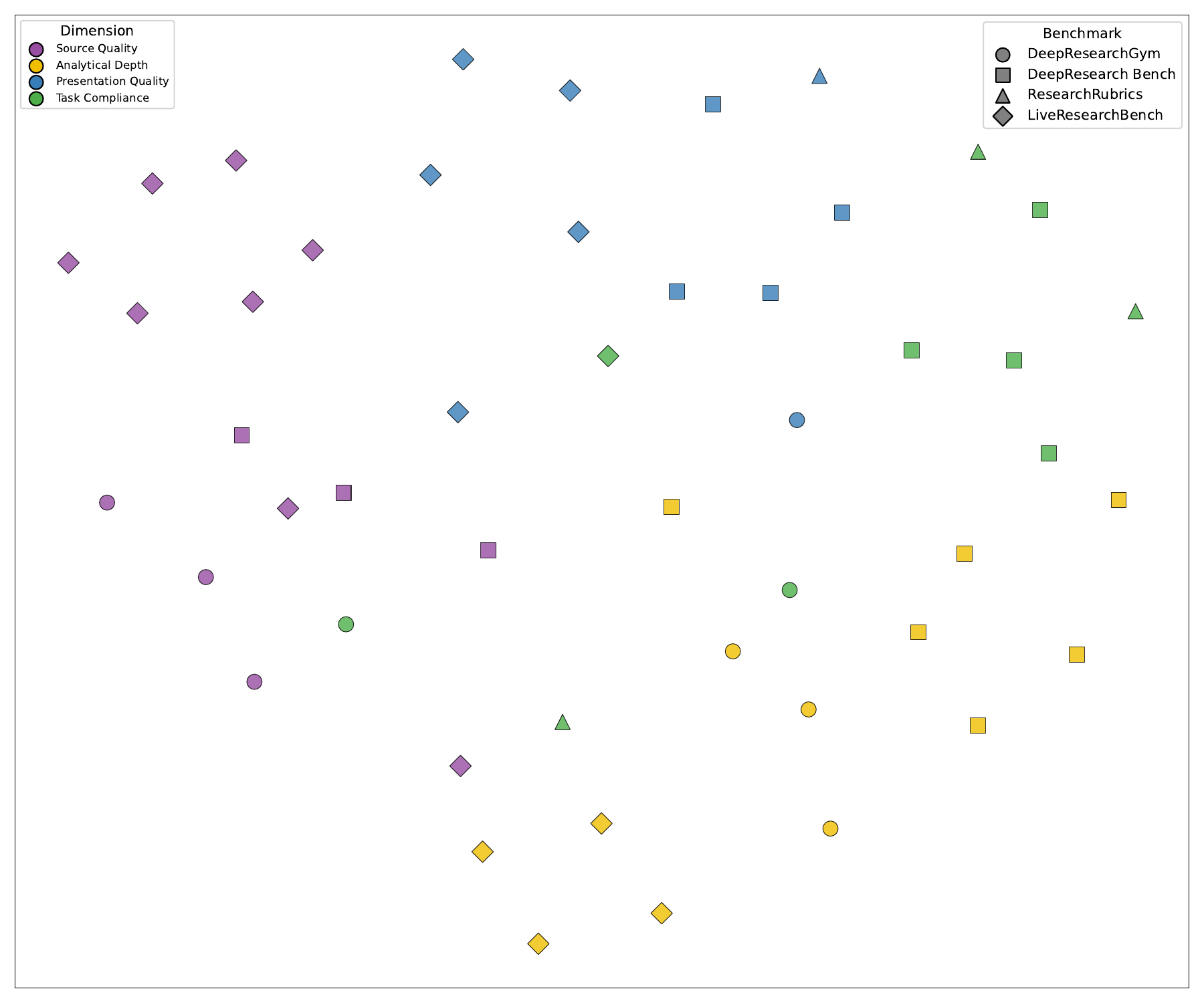}
    \caption{\textbf{Agentic Taxonomy Sub-Dimension Clustering Visualization.}
    Each point represents an evaluation criterion from one of four deep research benchmarks (shape), colored by its LLM-assigned taxonomy dimension.
    Criteria assigned to the same dimension tend to form localized regions in embedding space, suggesting coherent semantic groupings across the four
    verticals.}
    \label{fig:agentic_taxonomy}
    \vspace{-3ex}
\end{figure*}

As deep research evaluation gains momentum, multiple benchmarks have been proposed, each targeting overlapping yet distinct evaluation dimensions. While these efforts share common goals, differences in terminology, granularity, and evaluation focus make direct comparison difficult. To better understand the structure of existing evaluation practices, we develop an automated pipeline that organizes evaluation criteria from prior benchmarks into a unified, benchmark-agnostic taxonomy.

The pipeline consists of two stages. In the first stage, multiple LLM-based agents analyze each benchmark repository independently—DeepResearchGym~\citep{coelho2025deepresearchgym}, DeepResearch Bench~\citep{du2025deepresearch}, ResearchRubrics~\citep{sharma2025researchrubrics}, and LiveResearchBench~\citep{wang2025liveresearchbench}. Agents examine benchmark documentation, evaluation prompts, rubrics, and source code to extract leaf-level evaluation criteria alongside short natural language descriptions. Across the four benchmarks, this process yields 48 distinct criteria. All LLM-based stages in the pipeline use \texttt{Claude Opus 4.5} with fixed decoding settings (temperature=0) to ensure consistency across extraction and aggregation.

In the second stage, a single LLM clusters the extracted criteria into higher-level evaluation dimensions based on semantic similarity rather than benchmark provenance. To discourage benchmark-specific groupings, we impose a constraint that each resulting dimension must contain criteria originating from multiple benchmarks. We refer to individual extracted items as \emph{criteria}, which are grouped into \emph{dimensions} and summarized as a small set of high-level \emph{verticals}. This process reveals four recurring evaluation verticals observed across benchmarks:

\vspace{-1.5ex}
\begin{itemize}[leftmargin=*, itemsep=0.2pt]
    \item \textbf{Presentation Quality}: Writing clarity, structural organization, formatting consistency, and readability.
    \item \textbf{Task Compliance}: Adherence to instructions, coverage of requested topics, and fulfillment of explicit and implicit task requirements.
    \item \textbf{Analytical Depth}: Reasoning rigor, critical evaluation, originality of insights, and synthesis across multiple perspectives.
    \item \textbf{Source Quality}: Proper citation usage, source verification, and the extent to which claims are supported by credible, traceable evidence.
\end{itemize}

\vspace{-1ex}
To assess the semantic coherence of the induced taxonomy, we embed all extracted criteria using Cohere's \texttt{embed-english-v3.0}~\citep{cohere2023embedv3} model and visualize the resulting representations using UMAP (Figure~\ref{fig:agentic_taxonomy}). Criteria assigned to the same taxonomy dimension by the LLM tend to form localized regions in embedding space, providing qualitative evidence that the identified pillars correspond to coherent semantic groupings rather than artifacts of benchmark-specific phrasing.

\section{Metrics Details}\label{app:metrics_details}
In this section, we outline the implementation details and computation procedures for the suite of metrics employed in \AlgoName{}, categorized into Static Metrics (Writing Quality, Factuality, Citation Integrity, and Domain Authoritativeness) and Adaptive Metrics (Key-Information Coverage and Reasoning Quality).

\subsection{Writing Quality}
\label{app:wq_details}
Writing Quality (WQ) is a static, reference-free metric that evaluates the stylistic and structural presentation of the research report. It is assessed by an LLM Evaluator using a fixed rubric that scores the report along three weighted dimensions.

\begin{itemize}[leftmargin=*, itemsep=0.3pt]
    \item \textbf{Dimensions and Weights:}
    The score is computed as the average of three primary dimensions, each composed of weighted sub-dimensions, detailed in Table~\ref{tab:wq_rubric}:
    \begin{itemize}[leftmargin=*, itemsep=0.2pt]
        \item \textit{Ideas and Content} (33\%): Evaluates clarity of the main idea, relevance of details, information density, and conceptual synthesis.
        \item \textit{Organization} (33\%): Assesses heading structure, logical grouping of bullet points, and structural coherence.
        \item \textit{Sentence Fluency} (33\%): Measures rhythm/variety, transition smoothness, and readability/flow.
    \end{itemize}

    \item \textbf{Scoring Procedure:}
    For a given task $t$, the LLM Evaluator assigns a score $S_{t,d} \in [0, 100]$ for each dimension $d \in \{\text{Ideas and Content}, \text{Organization}, \text{Sentence Fluency}\}$ based on the sub-dimension weights. The final Writing Quality score for task $t$ is:
    \[
    \text{WQ}_t = \frac{1}{3} \sum_{d} S_{d,t}
    \]
\end{itemize}
\vspace{-2ex}

\begin{table*}[p]
\centering
\caption{\textbf{Writing Quality Rubric.} The exact descriptions used as system prompts for the LLM Evaluator to score each sub-dimension.}
\small
\renewcommand{\arraystretch}{1.4}
\begin{tabularx}{\textwidth}{l p{3.5cm} X}
\toprule
\textbf{Dimension} & \textbf{Sub-dimension (weight)} & \textbf{Evaluation Prompt (Exact Description)} \\
\midrule
\multirow{18}{*}{\textbf{Ideas and Content}} 
 & Main Idea Clarity (0.25) & This dimension assesses the clarity and specificity of the main idea expressed in the section summary. A high-quality section will present a \textbf{focused and well-articulated central idea} that is tightly aligned with the report question. Summaries lacking precision, or that simply list general topics without insight or framing, should be penalized. Do not give high scores if the main idea is vague, overgeneralized, or merely implied. \\
 & Detail Relevance (0.25) & This dimension focuses on how well the supporting details in the summary reinforce the main idea. Bullet points should be specific, relevant, and purposefully selected. Low-quality summaries may include off-topic, overly generic, or redundant details that do not support the section's main message. Do not reward high scores based on the amount of content alone—focus on alignment and purpose. \\
 & Information Density (0.25) & This dimension measures the \textbf{information richness} of the section summary. High-density summaries use each bullet to convey important, non-obvious, and topic-specific content. Shallow summaries repeat known facts, use vague language, or include fluff. Length alone should not be rewarded—focus on content value per line. \\
 & Conceptual Synthesis (0.25) & This dimension evaluates the \textbf{structural and conceptual integration} in the summary. Look for signs of synthesis such as: grouping related points, identifying contrasts, cause-effect relationships, or thematic framing. Poor summaries are unordered lists with no visible logic. Do not reward correctness alone—this dimension rewards insight, not just content. \\
\midrule
\multirow{12}{*}{\textbf{Organization}} 
 & Heading Structure (0.3) & This dimension assesses the use and clarity of headings in the section. High-quality summaries include headings that meaningfully segment the content, reflect topic hierarchy, and help orient the reader. Avoid rewarding default, generic, or misaligned headings. Headings should reflect actual conceptual boundaries. \\
 & Bullet Grouping Logic (0.4) & This dimension evaluates the internal logic of bullet groupings. High-quality summaries group related points together according to thematic, temporal, causal, or hierarchical logic. Low-quality groupings mix unrelated ideas, interrupt flow, or reflect no discernible principle. \\
 & Structural Coherence (0.3) & This dimension assesses whether the section's structure contributes to a logical, easy-to-follow reading experience. A coherent structure will show consistent flow from one part to the next, maintain logical transitions between bullet blocks, and avoid jarring shifts. Low-scoring sections often feel fragmented, with unclear order, repetition, or misplaced content. \\
\midrule
\multirow{14}{*}{\textbf{Sentence Fluency}} 
 & Rhythm \& Variety (0.3) & This dimension evaluates how naturally and dynamically the sentences flow. Strong writing features \textbf{variation in sentence length and structure}, avoiding repetitive patterns. Rhythm refers to the pacing and cadence of the prose—whether it reads with natural emphasis or becomes monotonous. High-scoring writing feels expressive and crafted, not just correct. \\
 & Transition Smoothness (0.3) & This dimension focuses on \textbf{how smoothly the sentences connect} to each other. High-quality prose includes \textbf{natural linking phrases, varied connectors, and logical sequencing}. Low-scoring writing jumps between ideas, or has jarring, abrupt shifts between sentences. Do not reward correctness alone—this dimension targets flow between thoughts. \\
 & Readability \& Flow (0.4) & This dimension evaluates the overall \textbf{readability and flow} of the paragraph. High-scoring writing reads smoothly aloud and requires little effort to follow. Low-scoring writing may include awkward phrasing, overcomplex or confusing sentence structures, or poor pacing. This metric captures the \textbf{global fluency} felt by readers, especially in multi-sentence passages. \\
\bottomrule
\end{tabularx}
\label{tab:wq_rubric}
\end{table*}

\subsection{Factuality}
\label{app:factuality_details}
Factuality evaluates whether the content of reports generated by the DRA is factually accurate with respect to external ground truth. The implementation utilizes a multi-stage verification pipeline that explicitly seeks both supporting and contradicting evidence for extracted claims.

\vspace{-1ex}
\begin{enumerate}[leftmargin=*, itemsep=0.3pt]
    \item \textbf{Key Factual Claim Extraction:} 
    An LLM extracts the $N=30$ most salient factual claims from the report with respect to the research question. The extraction is context-aware, incorporating the current date to resolve temporal references (e.g., ``current status'').
    \item \textbf{Neutralized Query Construction and Search:} 
    For each extracted claim, the system uses an LLM to generate multiple \textbf{neutralized} search queries. This step is critical to avoid \emph{confirmation bias}: rather than searching for the specific values or predicates in the claim (e.g., searching for ``inflation dropped to 2\%''), the system generates open-ended queries (e.g., ``current inflation rate'') to maximize the retrieval of both supporting and contradictory evidence.
    These queries are executed via a Web Search Tool, and results are deduplicated to form a pool of candidate source content.
    \item \textbf{Dual-Stream Evidence Extraction:} 
    Unlike standard RAG verification, we perform two distinct extraction passes on the retrieved content,
    \vspace{-0.5ex}
    \begin{itemize}
        \item \textit{Supporting Stream:} The LLM extracts passages that explicitly confirm the claim.
        \item \textit{Opposing Stream:} The LLM actively scans for passages that contradict or refute the claim.
    \end{itemize}
    \vspace{-0.5ex}
    \item \textbf{Factuality Judgment:} 
    The Judge LLM reviews the claim alongside the aggregated \textit{Supporting} and \textit{Opposing} evidence. It assigns one of four labels:
    \begin{itemize}[leftmargin=*, itemsep=0.3pt]
        \item \texttt{\textcolor{ForestGreen}{Supported}}: Evidence explicitly confirms the claim.
        \item \texttt{\textcolor{YellowGreen}{Partially Supported}}: Evidence supports some aspects of the claim but differs on details or is mixed.
        \item \texttt{\textcolor{BrickRed}{Contradicted}}: Evidence clearly refutes the claim.
        \item \texttt{\textcolor{YellowOrange}{Unverifiable}}: Evidence is insufficient, indirect, or too weak to make a reliable determination.
    \end{itemize}
\end{enumerate}
\paragraph{Factuality Scoring:} The Factuality score for task $t$, denoted as $\textrm{F}_t$, is calculated by weighting verified claims based on their support labels (1.0 point for \texttt{Support} and 0.5 point for \texttt{Partial Support}):
\[
    % F_t = \frac{N_{\ClaimSupport,t} + 0.5 N_{\ClaimPartSupport,t}}{N_{\ClaimSupport,t} + N_{\ClaimPartSupport,t} + N_{\ClaimContradict,t} + N_{\ClaimUnver,t}}
\textrm{F}_t = \frac{N_{\ClaimSupport,t} + 0.5 N_{\ClaimPartSupport,t}}{N_{\ClaimSupport,t} + N_{\ClaimPartSupport,t} + N_{\ClaimContradict,t}}\; .
\]
The overall Factuality score $F$ is computed as the average across all tasks in the dataset:
\[
    \textrm{F} = \frac{1}{|T|}\sum_{t\in T}{\textrm{F}_t}\; .
\]

\subsection{Citation Integrity}
\label{app:cf_details}
Citation Integrity (CI) evaluates the trustworthiness of a report by determining whether its claims are both explicitly attributed to citations and faithfully supported by the content of those sources. Relying on either metric in isolation is insufficient, as a report may achieve perfect citation coverage by hallucinating sources, or conversely, be factually accurate but lack transparency. We therefore decompose Citation Integrity into two constituent metrics---Claim Attribution (CA) and Citation Faithfulness (CF)---before unifying them into a single score.

The evaluation begins with a Verifiable Claim Extraction phase, where we identify all factual and argumentative assertions within the report. Crucially, an LLM filters out non-verifiable content—such as procedural meta-talk (e.g., ``The following section discusses...''), subjective commentary, and common knowledge—to establish a precise set of verifiable claims that forms the basis for evaluation.

\paragraph{Claim Attribution (CA). } This metric quantifies the extent to which the agent grounds its reasoning in external sources. It is defined as the ratio of distinct \emph{cited} claims to the total number of verifiable claims. A score of $1.0$ indicates that every verifiable claim is explicitly linked to a source URL, maximizing the report's auditability.

\paragraph{Citation Faithfulness (CF). } 
This metric evaluates the \textbf{veracity} of the provided evidence. For the subset of claims that are cited, CF determines whether the content of the source genuinely supports the claim. The workflow proceeds by retrieving the content of the associated URL for every cited claim and employing an LLM Judge to compare the text against the claim. The Judge assigns one of five labels: \texttt{\textcolor{ForestGreen}{Supported}}, \texttt{\textcolor{YellowGreen}{Partially Supported}}, \texttt{\textcolor{Orange}{Neutral}}, \texttt{\textcolor{BrickRed}{Contradicted}}, or \texttt{\textcolor{YellowOrange}{Unverifiable}}. The citation faithfulness score for task $t$, denoted $\CF_t$, is defined as:
\[
    \CF_t = \frac{N_{\ClaimSupport,t} + 0.5 N_{\ClaimPartSupport,t}}{N_{\ClaimSupport,t} + N_{\ClaimPartSupport,t} + N_{\textrm{neu},t} + N_{\ClaimContradict,t}}\; .
\]
The overall CF score is then computed as the average across tasks:
\[
    \CF = \frac{1}{|T|}\sum_{t\in T}{\CF_t}.
\]

\paragraph{Unified Citation Integrity Score.} To penalize trade-offs between attribution and faithfulness, we compute the final CI score as the harmonic mean of the two components:
\[
    \textrm{CI} = \frac{2 \cdot \textrm{CA} \cdot \CF}{\textrm{CA} + \CF}\; ,
\]
where $\textrm{CA}$ is the average claim attribution score across all tasks in the dataset. This formulation ensures that a high CI score is only achievable when an agent consistently supports its claims with valid, corroborating evidence; a failure in either (e.g., citing nothing, or citing everything incorrectly) will reduce the score.

\subsection{Domain Authoritativeness}\label{app:da_details}
Domain Authoritativeness (DA) evaluates the credibility and trustworthiness of the external sources cited by the agent. Unlike Citation Faithfulness, which checks if a specific URL supports a specific claim, DA assesses the reputation of the source domain itself, penalizing reliance on low-quality or unverifiable outlets (e.g., social media, clickbait farms) even if the content matches the claim.

The evaluation process proceeds as follows:

\begin{enumerate}[leftmargin=*, itemsep=0.3pt]
    \item \textbf{Domain Extraction and Deduplication:} 
    First, we aggregate all unique URLs cited across the report (derived from the CI pipeline). We extract the root domain from each URL (e.g., \texttt{https://www.nature.com/articles/xyz} $\rightarrow$ \texttt{nature.com}) and deduplicate them to ensure each source is evaluated only once per report.

    \item \textbf{LLM-Based Authority Assessment:} 
    An LLM Judge evaluates each unique domain using a rubric that assesses its overall reputation, credibility, and trustworthiness. The judge considers factors such as institutional backing, historical reliability, and editorial standards to classify the domain into a category (e.g., Government, Academic, News, Commercial) and assign an integer score $S_d \in [1, 10]$ based on the following scale:
    \begin{itemize}[leftmargin=*, itemsep=0.3pt]
        \item \textbf{Definitive Authority (9--10):} Gold-standard sources with institutional credibility (e.g., government agencies, top-tier academic institutions).
        \item \textbf{High Authority (7--8):} Trustworthy and credible sources (e.g., established news organizations).
        \item \textbf{Moderate Authority (4--6):} Acceptable but not ideal sources (e.g., general commercial sites).
        \item \textbf{Low Authority (1--3):} Unreliable sources with questionable credibility (e.g., social media platforms, unverified blogs).
    \end{itemize}

    \item \textbf{Scoring Formulation:}
    The scores are normalized to a $[0, 1]$ interval. The Domain Authoritativeness score for a task $t$, denoted as $\textrm{DA}_t$, is the average normalized score of all domains $D_t$ cited in that task:
    \[
    \textrm{DA}_t = \frac{1}{|D_t|} \sum_{d \in D_t} \frac{S_d}{10}
    \]
    The final metric is the average across all tasks in the dataset.
\end{enumerate}

\subsection{Key-Information Coverage}
\label{app:kic_details}
Key-Information Coverage (KIC) is an adaptive metric that measures whether the report addresses essential, query-specific facts. Unlike static comprehensiveness metrics, KIC uses an agent to retrieve up-to-date external knowledge to generate the evaluation criteria.

\vspace{-1ex}
\begin{itemize}[leftmargin=*, itemsep=0.6pt]
    \item \textbf{Protocol Creation (Adaptive):}
    For each query, the \textit{Protocol Creation Agent} (equipped with web search tools) identifies $K$ essential facts required for a complete answer. Each fact is converted into a yes/no question $q_k$ (e.g., ``Does the report mention the Jan 2026 deadline?'') grounded in retrieved evidence.

    \item \textbf{Evaluation:}
    The LLM Evaluator checks the generated report against each question $q_k$. Let $v_{k,t} \in \{0, 1\}$ be the verification result for question $k$ on task $t$.

    \item \textbf{KIC Scoring:}
    The score is the recall rate of these key facts:
    \vspace{-1ex}
    \[
    \text{KIC}_t = \frac{1}{K} \sum_{k=1}^{K} v_{k,t}
    \]
\end{itemize}
\vspace{-2ex}

\subsection{Reasoning Quality}
\label{app:rq_details}
Reasoning Quality (RQ) evaluates the logical rigor and analytical depth of the report. It focuses on the validity of arguments, the synthesis of disparate sources, and the avoidance of logical fallacies.
\vspace{-1ex}
\begin{itemize}[leftmargin=*, itemsep=0.3pt]
    \item \textbf{Validation Plan Generation:}
    The Protocol Creation Agent generates a set of open-ended, query-specific \textit{Challenging Questions}. For each question, it generates a structured \textit{Validation Plan} consisting of: (1) Extraction of reasoning chains from the report, (2) External verification using tools (e.g., web search, ArXiv, GitHub), and (3) Comparison criteria.

    \item \textbf{Agentic Execution:}
    The \textit{Agent Evaluator} executes the validation plan. It actively retrieves external evidence to verify the report's reasoning steps. Unlike KIC which checks for the presence of facts, RQ checks the soundness of the logic connecting them.

    \item \textbf{RQ Scoring:}
    The Agent Evaluator assigns a score $R_{t} \in [0, 10]$ based on a deductive rubric (starting at 10, deducting points for logical gaps, unsupported inferences, or ignored counter-evidence). The final normalized score is:
    \[
    \text{RQ}_t = \frac{R_{t}}{10}
    \]
\end{itemize}

\section{Controlled Experiments Details}\label{app:controlled_exp}
\subsection{Temporal Awareness in KIC}\label{subapp:temporal_awareness}

\begin{table}[b!]
    \vspace{-2ex}
    \centering
    \caption{\textbf{Temporal sensitivity analysis.} Comparison of DRB-RACE and DREAM-KIC scores across current (Upd.) and simulated outdated knowledge cutoffs (Jan 2025, Jan 2024) for the DRA. Unlike baselines, DREAM-KIC scores degrade on reports generated with older knowledge cutoffs, effectively capturing information lag.}    \label{tab:temporal_sensitivity}
    \fontsize{7pt}{8.5pt}\selectfont
    \resizebox{\textwidth}{!}{%
        \begin{tabular}{cl ccc ccc | ccc}
            \toprule
            \multirow{2}{*}{\textbf{ID}} & \multirow{2}{*}{\textbf{Topic}} & \multicolumn{3}{c}{\textbf{DRB-RACE Comp.}} & \multicolumn{3}{c}{\textbf{DRB-RACE Insight}} & \multicolumn{3}{c}{\textbf{DREAM-KIC}} \\
            \cmidrule(lr){3-5} \cmidrule(lr){6-8} \cmidrule(lr){9-11}
             & & Upd. & Jan-25 & Jan-24 & Upd. & Jan-25 & Jan-24 & Upd. & Jan-25 & Jan-24 \\
            \midrule
            1  & Current papal leadership \& policies & 48.29 & 53.80 & 55.75 & 48.55 & 52.12 & 55.23 & 64.00 & 0.00  & 0.00 \\
            2  & DOGE agency authority \& funding & 54.17 & 40.46 & 23.73 & 56.44 & 33.56 & 17.62 & 73.33 & 13.33 & 0.00 \\
            3  & US-China LLM competition (DeepSeek) & 50.62 & 47.10 & 44.75 & 50.14 & 45.90 & 39.76 & 84.38 & 56.13 & 31.25 \\
            4  & US-Venezuela military tensions & 46.96 & 47.07 & 49.15 & 44.38 & 44.32 & 47.28 & 46.88 & 6.12  & 0.00 \\
            5  & Sudan civil war status & 48.32 & 48.72 & 0.00  & 46.21 & 48.46 & 0.00  & 81.67 & 52.00 & 0.00 \\
            6  & UK Prime Minister's challenges & 55.57 & 50.59 & 48.58 & 55.64 & 48.17 & 42.11 & 64.00 & 43.80 & 36.20 \\
            7  & Boeing CEO \& Starliner mission & 50.03 & 51.77 & 49.86 & 49.87 & 52.62 & 44.32 & 94.44 & 78.00 & 5.33 \\
            8  & Nvidia market cap ranking & 52.30 & 50.26 & 49.29 & 48.41 & 47.49 & 42.67 & 78.00 & 38.89 & 22.44 \\
            9  & Latest Best Picture winner & 51.44 & 48.22 & 51.18 & 48.60 & 46.31 & 51.08 & 84.87 & 6.13  & 6.13 \\
            10 & Ukrainian Kursk occupation & 50.28 & 46.63 & 40.69 & 48.43 & 44.93 & 40.43 & 80.00 & 43.33 & 16.67 \\
            11 & Protein structure prediction SOTA & 48.82 & 49.73 & 45.86 & 44.78 & 49.05 & 41.73 & 65.75 & 71.88 & 18.88 \\
            12 & US Strategic Bitcoin Reserve & 48.34 & 79.21 & 82.53 & 52.59 & 73.29 & 60.24 & 81.33 & 33.00 & 11.44 \\
            13 & Digital Euro implementation & 51.44 & 42.92 & 41.17 & 56.50 & 42.30 & 38.28 & 97.00 & 25.00 & 21.75 \\
            14 & Tesla Cybercab deployment & 46.77 & 44.88 & 40.64 & 53.36 & 45.73 & 39.58 & 75.00 & 60.71 & 39.14 \\
            15 & GLP-1 drug supply status & 51.09 & 47.05 & 42.01 & 50.07 & 49.30 & 41.09 & 73.70 & 82.78 & 56.13 \\
            16 & TikTok divestiture deadline & 48.61 & 51.67 & 43.71 & 53.70 & 60.86 & 44.90 & 95.83 & 50.00 & 16.58 \\
            17 & SpaceX Starship IFT results & 47.47 & 47.86 & 47.46 & 51.22 & 48.01 & 45.85 & 81.25 & 49.62 & 34.25 \\
            18 & Taylor Swift Eras Tour impact & 50.62 & 47.21 & 43.18 & 53.06 & 47.71 & 45.78 & 92.20 & 60.20 & 56.00 \\
            19 & Paramount-Skydance merger status & 50.87 & 51.61 & 47.71 & 48.40 & 40.44 & 51.33 & 73.33 & 36.67 & 10.00 \\
            20 & Disney CEO succession & 48.32 & 53.94 & 48.89 & 50.51 & 55.10 & 44.31 & 100.00 & 88.35 & 64.53 \\
            \midrule
            & \textbf{AVG} & \textbf{50.02} & \textbf{50.04} & \textbf{44.81} & \textbf{50.54} & \textbf{48.78} & \textbf{41.68} & \textbf{79.35} & \textbf{44.80} & \textbf{22.34} \\
            \bottomrule
        \end{tabular}%
    }
\end{table}

To generate the reference reports for RACE, we utilized Smolagents' open Deep Research agent~\citep{roucher2025open}, an open-source CodeAgent-based framework. We performed three independent generation runs per query to create distinct report versions: one reflecting current information, and two simulating outdated knowledge bases with cutoffs set to January 1st, 2025, and January 1st, 2024, respectively. These simulated cutoffs were enforced by appending specific temporal constraints to the system prompt and strictly filtering results from the web search tool to exclude any content published after the target dates. 

In \Cref{tab:temporal_sensitivity}, we present the complete evaluation results for DRB--RACE Comprehensiveness, Insight, and DREAM--KIC across all 20 queries. The data indicates that DRB--RACE lacks consistent temporal sensitivity. While Comprehensiveness scores do drop on average for the Jan-2024 cutoff (e.g., Query 2 ``DOGE agency authority''), the metric is effectively blind to the Jan-2025 cutoff, with average scores remaining virtually identical to the baseline. Furthermore, RACE metrics frequently exhibit paradoxical behavior, where scores increase for outdated reports (e.g., Query 1 ``Current papal leadership''). In contrast, DREAM--KIC demonstrates a far stronger alignment with the information lag. Although not strictly monotonic (scores fluctuated in two instances -- Queries 11 and 15) DREAM--KIC typically imposes drastic penalties as the knowledge horizon recedes, effectively capturing the absence of specific, recent developments that the baseline metrics miss.

\begin{table*}[th!]
\centering
\caption{\textbf{Task Alignment Criteria Comparison.} DeepResearch Bench RACE rubric criteria alongside DREAM KIC Checklist items for the query: ``Prepare a regulatory compliance update on the legal status of TikTok in the U.S. Specifically, clarify the proximity of the deadline for ByteDance to divest its U.S. operations to avoid a nationwide app store ban.''}
\fontsize{7pt}{8.5pt}\selectfont
\setlength{\tabcolsep}{3pt}
\renewcommand{\arraystretch}{1.2}
\begin{tabular}{c|p{0.63\textwidth}|p{0.31\textwidth}}
\toprule
\textbf{\#} & \textbf{RACE Criteria} & \textbf{DREAM KIC Checklist Items} \\
\midrule
& \cellcolor{gray!10}\textit{Comprehensiveness} & \\
\midrule
1 & \textbf{Identification and Explanation of Core Legislation}: Assesses if the article clearly identifies the specific U.S. law mandating the potential ban (e.g., the `Protecting Americans from Foreign Adversary Controlled Applications Act') and explains its key provisions. This is the foundational context for the entire update. 
& Does the report mention the Protecting Americans from Foreign Adversary Controlled Applications Act (PAFACA) as the specific law requiring ByteDance to divest TikTok? \\
\midrule
2 & \textbf{Detailed Breakdown of the Divestiture Deadline and Extensions}: Evaluates if the article provides the exact initial deadline for divestiture and, crucially, clarifies the conditions and duration of any potential presidential extensions. This directly addresses the task's specific requirement to clarify the deadline's `proximity.' 
& Does the report state that the current divestiture deadline is January 23, 2026? \\
\midrule
3 & \textbf{Coverage of Ongoing Legal Challenges}: Checks if the article discusses the lawsuits filed by TikTok/ByteDance or other parties to challenge the law. This is essential for a comprehensive understanding of the current `legal status,' as these proceedings could impact the deadline and enforcement. 
& Does the report accurately calculate and convey that approximately 31 days or one month remains until the deadline from late December 2025? \\
\midrule
4 & \textbf{Analysis of Key Stakeholder Positions}: Assesses whether the report covers the stances and actions of key stakeholders, including the U.S. Executive and Legislative branches, ByteDance, and the Chinese government (regarding its approval of tech exports). This provides a complete picture of the geopolitical and corporate landscape. 
& Does the report mention that ByteDance signed a deal in mid-December 2025 to create a U.S. joint venture, with closing scheduled for January 22, 2026? \\
\midrule
5 & \textbf{Explanation of the Ban's Enforcement Mechanism}: Evaluates if the article explains how the ban would be implemented—specifically, the nature of the nationwide app store ban and its implications for app distribution and hosting services. This clarifies the practical consequences of non-compliance. 
& Does the report specify the ownership structure of the joint venture deal, including that ByteDance will retain 19.9\% and U.S. investors will hold majority control? \\
\midrule
6 & \textbf{Inclusion of Relevant Historical Context}: Checks for a brief summary of previous U.S. government actions or attempts to regulate TikTok (e.g., under the Trump administration). This provides depth and demonstrates a thorough understanding of the long-standing nature of the issue. 
& Does the report mention that the Supreme Court upheld the TikTok divestiture law in January 2025? \\
\midrule
& \cellcolor{gray!10}\textit{Insight} & \\
\midrule
7 & \textbf{Analysis of Legal Challenges and Timeline Impact}: Evaluates if the article moves beyond merely stating that lawsuits exist to deeply analyze the core legal arguments (e.g., First Amendment rights), assess their potential to secure an injunction or stay, and logically connect these legal outcomes to the potential alteration of the divestiture deadline. This is the primary factor defining the deadline's stability. 
& Does the report mention that the original deadline was January 19, 2025, and that TikTok briefly went dark on that date? \\
\midrule
8 & \textbf{Synthesis of Factors into Forward-Looking Scenarios}: Assesses the ability to integrate the legal, political, and business analyses into a coherent forecast of potential outcomes (e.g., successful sale, ban proceeds after court losses, extended legal stalemate). This demonstrates a holistic understanding and provides the most valuable strategic insight. 
& Does the report document that President Trump granted multiple extensions to the TikTok deadline throughout 2025? \\
\midrule
9 & \textbf{Depth of Divestiture Feasibility Assessment}: Evaluates the analysis of the practical barriers to a sale, such as the technical complexity of separating TikTok's recommendation algorithm, the financial valuation challenges, and the need for approval from the Chinese government. This insight is crucial for determining if the deadline is realistically achievable. 
& Does the report explain the specific consequences of missing the deadline, including prevention of app store updates and distribution? \\
\midrule
10 & \textbf{Contextualization of the Deadline's Political Significance}: Assesses if the article effectively frames the deadline within the broader U.S.-China geopolitical tensions and U.S. domestic political motivations. This provides insight into the `why' behind the law and the political will to enforce it, which influences any potential for extensions or negotiated solutions. 
& Does the report mention the bipartisan congressional support for the TikTok divestiture law? \\
\midrule
11 & \textbf{Identification of Key Signposts for Future Monitoring}: Evaluates whether the analysis identifies specific, critical future events or indicators (e.g., key court dates, statements from potential buyers, Chinese regulatory responses) that readers should watch to anticipate how the situation will evolve. This transforms the analysis into an actionable intelligence tool. 
& Does the report explain the national security concerns about data sharing with China that motivated the law? \\
\midrule
12 & & Does the report mention that TikTok has approximately 170 million U.S. users? \\
\midrule
13 & & Does the report mention the valuation of TikTok's U.S. operations in the context of the deal? \\
\midrule
14 & & Does the report acknowledge that the deal requires Chinese regulatory approval, creating potential uncertainty? \\
\bottomrule
\end{tabular}
\label{tab:race_kic_tiktok_comparison}
\end{table*}
To complement the results in \Cref{fig:timeaware_example}, the complete set of DRB--RACE Comprehensiveness and Insight criteria as well as KIC checklist items is provided in \Cref{tab:race_kic_tiktok_comparison}.

\subsection{Detecting Reasoning Flaws}\label{subapp:reasoning}
% \begin{figure*}[t]
%     \centering
%     \includegraphics[width=1.0\textwidth]{figures/rqe_example}
%     \caption{\textbf{Reasoning Quality Evaluation Example.}
%     Example for reasoning quality evaluation criterion. The agent evaluator detects flaws in the trip planned by the DRA.}
%     \label{fig:agentic_taxonomy}
% \end{figure*}

We curated ten research queries spanning domains where analytical reasoning is critical, including policy analysis, technical system comparisons, and causal explanations. Each query was designed to require multi-step reasoning and evidence synthesis, making them suitable for evaluating reasoning quality detection. The complete query set is provided in Table~\ref{tab:reasoning_queries}. To generate the reports, we used Smolagents' open Deep Research agent~\citep{roucher2025open}. For each query, we generated two report variants using identical source materials. High-quality reports were generated following standard analytical practices with sound logical structure, properly supported claims, and coherent argumentation. Malformed reports were systematically injected with reasoning flaws while preserving surface-level fluency and plausibility; injected flaws included unsupported causal claims, circular reasoning, false equivalences, and cherry-picked evidence. This controlled design yielded 20 total reports (10 high-quality, 10 malformed) forming 10 matched pairs, enabling direct comparison while controlling for query-specific factors. Each report was evaluated using both DREAM--RQ and DRB--RACE frameworks.

\begin{table}[t]
\centering
\caption{Research queries for the ``Reasoning Flaws Detection'' experiment in \Cref{sec:reasoning}.}
\small
\begin{tabularx}{\textwidth}{@{}c X@{}} 
\toprule
\textbf{ID} & \textbf{Query} \\
\midrule
1 & How will a 10\% price cut affect unit sales next month for an e-commerce product line? \\ 
\midrule
2 & What assumptions about inflation, interest rates, and currency stability most influence predictions of stock market volatility in emerging economies over the next five years? \\ 
\midrule
3 & How will the adoption of renewable energy technologies impact global oil demand over the next decade? \\ 
\midrule
4 & What assumptions about hospital staffing, patient flow, and technology adoption most influence predictions of emergency department wait times over the next five years? \\ 
\midrule
5 & How will AI-powered recommendation engines affect consumer purchase diversity in online retail markets over the next two years? \\ 
\midrule
6 & What assumptions about monetary policy, housing supply, and demographic trends most influence predictions of urban housing prices over the next decade? \\ 
\midrule
7 & How will climate change affect agricultural crop yields in major exporting countries over the next 20 years? \\ 
\midrule
8 & What assumptions about vaccine distribution, mutation rates, and healthcare access most influence predictions of pandemic recovery times? \\ 
\midrule
9 & How will quantum computing adoption influence financial risk modeling practices over the next 15 years? \\ 
\midrule
10 & How would a heatwave next month affect ice cream demand? \\
\bottomrule
\end{tabularx}
\label{tab:reasoning_queries}
\end{table}

% \begin{table}[htbp]
% \centering
% \small
% \begin{tabular}{p{13cm}}
% \toprule
% Query \\
% \midrule
% How will a 10\% price cut affect unit sales next month for an e-commerce product line? \\
% What assumptions about inflation, interest rates, and currency stability most influence predictions of stock market volatility in emerging economies over the next five years? \\
% How will the adoption of renewable energy technologies impact global oil demand over the next decade? \\
% What assumptions about hospital staffing, patient flow, and technology adoption most influence predictions of emergency department wait times over the next five years? \\
% How will AI-powered recommendation engines affect consumer purchase diversity in online retail markets over the next two years? \\
% What assumptions about monetary policy, housing supply, and demographic trends most influence predictions of urban housing prices over the next decade? \\
% How will climate change affect agricultural crop yields in major exporting countries over the next 20 years? \\
% What assumptions about vaccine distribution, mutation rates, and healthcare access most influence predictions of pandemic recovery times? \\
% How will quantum computing adoption influence financial risk modeling practices over the next 15 years? \\
% How would a heatwave next month affect ice cream demand? \\
% \bottomrule
% \end{tabular}
% \caption{Research Queries for Reasoning Flaw Detection Experiment}
% \label{tab:reasoning_queries}
% \end{table}

\subsection{Grounding Beyond Citation Faithfulness}\label{subapp:factuality_beyond_CF}
To rigorously test the hypothesis that citation-alignment metrics are blind to well-cited falsehoods, we constructed a controlled dataset focusing on \emph{extrinsic} factual errors, that is, claims that are false in reality but are accompanied by a supporting citation. We manually curated 15 adversarial claim pairs. Each pair consists of:
\begin{itemize}
    \item A \textbf{Ground Truth} variant ($c_{\text{true}}$) accompanied by a valid source URL.
    \item A \textbf{Plausible Hallucination} variant ($c_{\text{false}}$) accompanied by a matching, misleading URL.
\end{itemize}
Crucially, the matching URLs for the false variants were selected to satisfy standard citation faithfulness checks, serving as specious support for the incorrect claims. These sources span various modes of invalidity, such as outdated information—for instance, citing a 2019 study claiming 46\% of Bitcoin transactions are illicit, which directly contradicts current data showing less than 1\%. Other examples leverage persistent myths, such as the mechanic's rule of thumb to change oil every 3,000 miles despite modern engineering standards, or fringe narratives, like articles framing 15-minute cities as government-imposed lockdown zones. This setup ensures that the falsity stems from the \emph{content} of the claim relative to objective reality, rather than a mismatch between the claim and its provided citation. The complete set of pairs, contrasting ground truth with aligned misinformation, is listed in \Cref{tab:factuality_corruption_examples}.

\begin{table*}[htbp]
\centering
\caption{\textbf{Adversarial Factuality Dataset.} The 15 adversarial claim pairs used to measure factual sensitivity. Ground Truth (\textcolor{ForestGreen}{T}) variants are backed by reliable evidence, while False (\textcolor{BrickRed}{F}) variants are paired with specious citations that satisfy text-alignment criteria but contradict objective reality.}
\fontsize{7.6}{9}\selectfont
\renewcommand{\arraystretch}{1.25}
\begin{tabularx}{\textwidth}{c l c X l}
\toprule
\textbf{ID} & \textbf{Topic} & \textbf{Variant} & \textbf{Claim} & \textbf{URL Domain} \\
\midrule
\multirow{4}{*}{1} & \multirow{4}{*}{15-Min Cities} & \textcolor{ForestGreen}{T} & 15-minute cities are an urban planning concept to place essential services within walking distance; they do not restrict movement. & fullfact \\
& & \textcolor{BrickRed}{F} & The `15-minute city' is a lockdown plan to restrict residents' movement and fine them for leaving their zones. & spiked-online \\
\midrule
\multirow{4}{*}{2} & \multirow{4}{*}{GMO Safety} & \textcolor{ForestGreen}{T} & Commercially available GMO foods are safe for consumption and do not pose health risks. & natl-academies \\
& & \textcolor{BrickRed}{F} & There is no scientific consensus on GMO safety; they pose potential long-term risks to health and biodiversity. & ensser \\
\midrule
\multirow{4}{*}{3} & \multirow{4}{*}{Climate Change} & \textcolor{ForestGreen}{T} & Rapid warming since the mid-20th century is primarily driven by human activities, specifically fossil fuel emissions. & nasa \\
& & \textcolor{BrickRed}{F} & Climate forecasting has fundamental uncertainties regarding natural variability; warming may not be as catastrophic as models predict. & judithcurry \\
\midrule
\multirow{3}{*}{4} & \multirow{3}{*}{Homeopathy} & \textcolor{ForestGreen}{T} & Homeopathy is a pseudoscience with no medicinal effect beyond placebo. & nhs \\
& & \textcolor{BrickRed}{F} & Homeopathy is a clinically effective medical treatment for various conditions beyond the placebo effect. & hri-research \\
\midrule
\multirow{4}{*}{5} & \multirow{4}{*}{Dino--Human} & \textcolor{ForestGreen}{T} & Non-avian dinosaurs went extinct approximately 65 million years before modern humans evolved. & usgs \\
& & \textcolor{BrickRed}{F} & Humans and dinosaurs lived on Earth at the same time, approximately 6,000 years ago. & answersingenesis \\
\midrule
\multirow{4}{*}{6} & \multirow{4}{*}{Moon Landing} & \textcolor{ForestGreen}{T} & On July 20, 1969, Apollo 11 landed on the Moon; Neil Armstrong and Buzz Aldrin became the first humans to walk on its surface. & britannica \\
& & \textcolor{BrickRed}{F} & The 1969 Apollo moon landings were staged events, as evidenced by photographic anomalies. & time \\
\midrule
\multirow{3}{*}{7} & \multirow{3}{*}{5G Health} & \textcolor{ForestGreen}{T} & 5G frequencies are non-ionizing and safe for humans within international guidelines. & who \\
& & \textcolor{BrickRed}{F} & 5G technology will massively increase exposure to radiofrequency radiation, which is proven harmful to humans. & ehtrust \\
\midrule
\multirow{4}{*}{8} & \multirow{4}{*}{Bermuda Triangle} & \textcolor{ForestGreen}{T} & There is no evidence that disappearances occur more frequently in the Bermuda Triangle than in other well-traveled ocean areas. & noaa \\
& & \textcolor{BrickRed}{F} & The Bermuda Triangle is a deadly zone where ships and planes disappear at a frequency far exceeding statistical probability. & smu \\
\midrule
\multirow{4}{*}{9} & \multirow{4}{*}{Knuckle Cracking} & \textcolor{ForestGreen}{T} & Knuckle cracking creates a popping sound due to cavitation; there are no known detrimental effects or links to arthritis. & houstonmethodist \\
& & \textcolor{BrickRed}{F} & Cracking knuckles consistently can wear away cartilage and increase risk of developing arthritis. & nih \\
\midrule
\multirow{2}{*}{10} & \multirow{2}{*}{10\% Brain} & \textcolor{ForestGreen}{T} & Humans use virtually 100\% of their brains throughout the day, even during sleep. & britannica \\
& & \textcolor{BrickRed}{F} & Most humans only use 10\% of their brain capacity, leaving vast potential untapped. & gaia \\
\midrule
\multirow{3}{*}{11} & \multirow{3}{*}{Google Energy} & \textcolor{ForestGreen}{T} & A Google search uses about 0.0003 kWh, orders of magnitude less energy than boiling a kettle. & google \\
& & \textcolor{BrickRed}{F} & Two Google searches generate the same amount of CO2 as boiling a kettle for tea. & hindustantimes \\
\midrule
\multirow{4}{*}{12} & \multirow{4}{*}{EV Emissions} & \textcolor{ForestGreen}{T} & Even accounting for battery manufacturing, total lifetime GHG emissions of an EV are lower than a comparable gasoline car. & epa \\
& & \textcolor{BrickRed}{F} & Widespread EV adoption may increase emissions due to energy-intensive mining and refining of battery materials. & manhattan-inst \\
\midrule
\multirow{3}{*}{13} & \multirow{3}{*}{Crypto Illicit} & \textcolor{ForestGreen}{T} & Illicit activity accounts for less than 1\% of total cryptocurrency transaction volume. & chainalysis \\
& & \textcolor{BrickRed}{F} & About 46\% of bitcoin transactions are involved in illegal activity, transforming black markets. & repec \\
\midrule
\multirow{4}{*}{14} & \multirow{4}{*}{Earth Shape} & \textcolor{ForestGreen}{T} & The Earth is an oblate spheroid, confirmed by satellite imagery, gravity, and centuries of astronomical observation. & wikipedia \\
& & \textcolor{BrickRed}{F} & The Earth is a stationary plane; the 'South Pole' is an impenetrable ice wall surrounding the world. & gutenberg \\
\midrule
\multirow{4}{*}{15} & \multirow{4}{*}{Oil Change} & \textcolor{ForestGreen}{T} & Modern vehicles using synthetic oils can go 7,500--10,000 miles between changes, far surpassing the 3,000-mile guideline. & kbb \\
& & \textcolor{BrickRed}{F} & Engine oil must be changed every 3,000 miles because additives break down, causing sludge and reducing lubrication. & artmorse \\
\bottomrule
\end{tabularx}
\label{tab:factuality_corruption_examples}
\end{table*}

To isolate the sensitivity of the metrics, we bypassed the claim-source extraction phase and directly constructed synthetic evaluation payloads. For a given corruption level $r \in [0, 1]$,  we assembled a test batch of size $N=15$ by selecting $r \times N$ false variants and $(1-r) \times N$ true variants. These batches were then fed directly into the evaluation pipelines of both DRB--FACT and \AlgoName{}--Factuality.

The divergence in performance, clearly visualized in \Cref{fig:teaser} (middle), highlights a fundamental difference in evaluation scope. DRB--FACT proved insensitive to the corruption, with citation accuracy scores remaining effectively constant ($\approx 100\%$) across the entire sweep. Restricted to verifying consistency with the provided URL, the metric correctly identified that the false claims matched their sources, but failed to detect the underlying misinformation. In sharp contrast, \AlgoName{}--Factuality exhibited a near-linear degradation inversely proportional to the corruption rate $r$. By independently retrieving fresh evidence from the web, the evaluator successfully identified the external contradictions for the false variants, correctly rejecting the misleading citations provided in the input.

\section{Additional Details of Human Evaluation}
\label{app:human_eval}

\subsection{Human Evaluation Effort} \label{subsec: human evaluation details}
The evaluation process entails a rigorous workflow to ensure reliable consistency. For each task within DeepResearch Bench~\cite{du2025deepresearch}, the core data points expected in the report were first identified, followed by a comprehensive review of the three corresponding protocols. This substantial effort ensures that the resulting feedback is sufficiently detailed to serve as a robust ground truth for calculating precision and recall against human judgments.

\subsection{Protocol Creation Methods} \label{subsec:recipe_generation_methods}
To assess the effectiveness of our approach, we compare three different strategies for creating evaluation protocols:

\begin{itemize}[leftmargin=*]
    \item \textbf{Direct LLM generation:} The LLM is directly prompted with the task description and produces evaluation criteria in a single step. It relies solely on its internal knowledge and has no access to external resources.
    \item \textbf{Agent without external knowledge:} An agentic system generates evaluation criteria through multi-step reasoning. Although the agent does not have access to external knowledge, the multi-step process enables deeper analysis compared to direct LLM generation.
    \item \textbf{Agent with external knowledge (\AlgoName{}):} Our final setting equips the agent with both multi-step reasoning and access to external knowledge capabilities. The agent can query external sources (e.g., web search, and ArXiv/GitHub for domain-specific knowledge) to supplement its analysis.
\end{itemize}

% \begin{table}[t]
% \centering
% \small
% \begin{tabularx}{\linewidth}{l X}
% \toprule
% \textbf{Dimension} & \textbf{Scoring Guidelines} \\
% \midrule
% \textit{Relevance} & The evaluation point directly targets the key aspects of the original query.  \\
% \midrule
% \textit{Verifiability}
%  & The evaluation point can be verified using accessible sources or well-defined methods. \\
% \midrule
% \textit{Clarity}
%  & The evaluation point is precise, unambiguous, and easy to understand. \\
% \bottomrule
% \end{tabularx}
% \caption{\textbf{Human evaluation rubric}. Criteria used to guide expert assessment of created protocols. }
% \label{tab:prime_human_evaluation_rubric}
% \end{table}

\begin{table}[b]
\centering
\caption{\textbf{Human evaluation rubric}. Criteria used to guide expert assessment of created protocols.}
% \small
\begin{tabularx}{\linewidth}{l X}
\toprule
\textbf{Dimension} & \textbf{Scoring Guidelines} \\
\midrule
\textit{Relevance} & Directly addresses the specific requirements and key aspects of the research query. \\
\midrule
\textit{Verifiability}
 & Can be objectively confirmed using accessible external evidence or well-defined logic. \\
\midrule
\textit{Clarity}
 & Is formulated precisely and unambiguously to ensure consistent interpretation. \\
\midrule
\textit{Validation}
 & Is methodologically correct, free of circular reasoning, and capable of rigorously confirming the answer. \\
\bottomrule
\end{tabularx}
\label{tab:prime_human_evaluation_rubric}
\end{table}
\subsection{Human Evaluation Protocol} \label{subsec: evaluation details}
Human evaluators are instructed to assess each evaluation point in the generated protocols based on relevance, clarity, and verifiability, with validation soundness additionally included for RQ. The detailed definitions of these criteria are provided in Table~\ref{tab:prime_human_evaluation_rubric}. To ensure a fair comparison, protocols produced under the three different strategies are evaluated using the same criteria. Each evaluation point is rated on a 1–3 scale (where higher scores indicate better quality), which is then normalized to a 0–1 scale. We further analyze inter-annotator agreement across evaluators in Section~\ref{subsec: agreement analysis} to confirm the reliability of the human assessment.

\subsection{Inter-Annotator Agreement Analysis} \label{subsec: agreement analysis}
We compute Kendall’s coefficient of concordance (Kendall’s W) to measure annotator agreement on protocol quality. Kendall’s W quantifies the level of concordance among raters when assessments are ranked; here, it reflects the degree to which annotators share similar preferences across the different creation strategies. As reported in Table~\ref{tab:human_eval_agreement}, annotators exhibit moderate agreement on Relevance and Clarity ($W \approx 0.58$), and substantial agreement on Verifiability and Validation Soundness ($W > 0.75$). The overall agreement is statistically significant across all indicators.

We attribute the variance in agreement primarily to the similarity between the \textit{Direct LLM} and \textit{Agent without retrieval} baselines. Since both rely solely on the same backbone model's internal knowledge, their outputs are often qualitatively similar, making it difficult for evaluators to consistently rank one over the other. However, the \textit{Agent with retrieval} is consistently distinguished from these baselines, confirming that the human preference for retrieval-augmented protocols is reliable.

\begin{table}[H]
\centering
\caption{Agreement analysis among human annotators, measured using Kendall’s W. All results are statistically significant ($p < .05$).}
\begin{tabular}{lcc}
\toprule
Indicator & Kendall’s W & p-value \\
\midrule
Relevance            & 0.58 & .030 \\
Verifiability        & 0.78 & .009 \\
Clarity              & 0.58 & .030 \\
Validation Soundness & 0.86 & .006 \\
\midrule
\textbf{Average}     & \textbf{0.69} & \textbf{.016} \\
\bottomrule
\end{tabular}
\label{tab:human_eval_agreement}
\end{table}
% \section{Additional Details of Human Evaluation}
% \label{app:human_eval}

\section{Detailed DREAM Benchmarking Results}\label{sec:dream_benchmarking}
In this section, we present complete evaluation results for three Deep Research Agents (DRAs) across our three benchmarks, complementing the results in \Cref{sec:benchmarking}. A key advantage of the \AlgoName{} framework is its dataset-agnostic design, which enables us to normalize scores across heterogeneous tasks and compute a unified aggregate performance profile.

\paragraph{Performance Overview.} Table~\ref{tab:dra_results} details the performance of each system on static metrics (WQ, Factuality, CI, DA) and adaptive metrics (KIC, RQ), alongside a composite score averaging results across all metrics. As discussed in \Cref{sec:benchmarking}, distinct system profiles emerge: Smolagents Open DR leads in most metrics (except CI), though behavior may depend on the dataset, e.g., on \textsc{DeepResearchBench}, Tongyi Deep Research has higher WQ and Factuality.

\begin{figure*}[h]
    \centering
    \includegraphics[width=\textwidth]{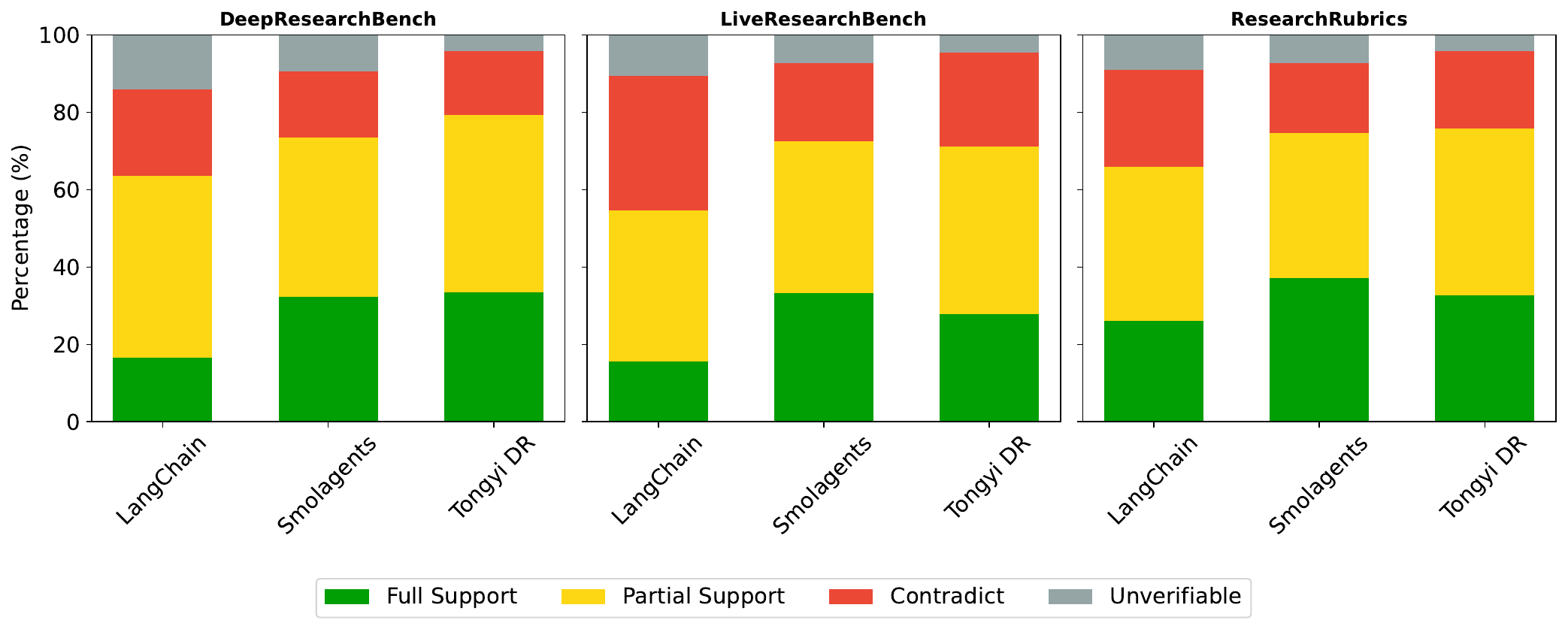}
    \caption{\textbf{Factuality Label Distribution.} Distribution of factuality judgments (Full Support, Partial Support, Contradict, Unverifiable) for each model across three datasets.}    \label{fig:factuality_breakdown}
\end{figure*}

\paragraph{Source Quality Analysis.} The detailed breakdown of our Source Quality-related metrics reveals clear distinctions in verification capabilities. Figure~\ref{fig:factuality_breakdown} presents the distribution of Factuality labels across all three benchmarks. Smolagents typically achieves higher proportion of ``Full Support'' judgments (green bars), indicating a superior ability to generate claims that are independently corroborated by retrieved external evidence. In contrast, LangChain exhibits higher rates of ``Unverifiable'' or `Contradict' claims.

\begin{figure*}
    \centering
    \includegraphics[width=\textwidth]{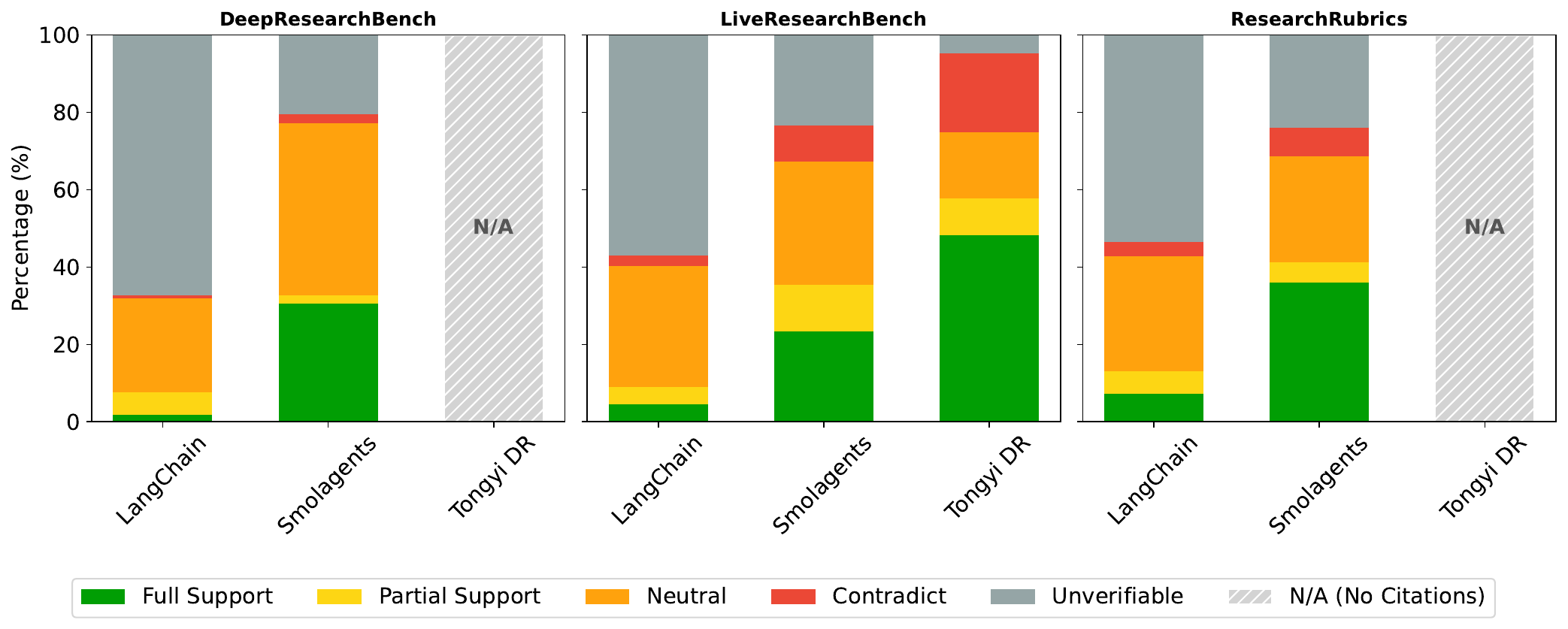}
    \caption{\textbf{Citation Faithfulness Label Distribution.} Distribution of citation faithfulness labels assessing the alignment between claims and cited source text across three datasets.}
    \label{fig:faithfulness_breakdown}
\end{figure*}

\begin{figure*}
    \centering
    \includegraphics[width=.9\textwidth]{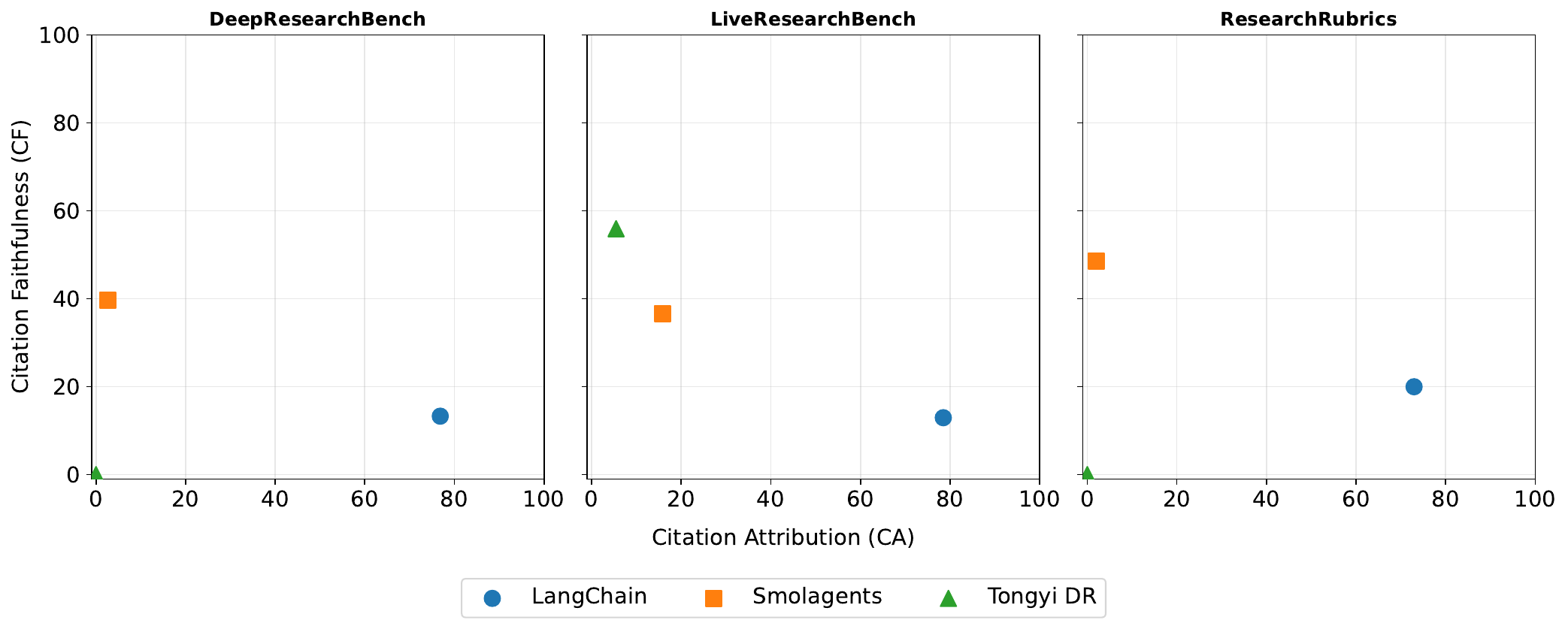}
    \caption{\textbf{Citation Integrity Components.} Citation Faithfulness and Citation Attribution visualized for each model across three benchmarks.}
    \label{fig:ci_scatter}
\end{figure*}

A similar pattern emerges in Figure~\ref{fig:faithfulness_breakdown},
which details the Citation Faithfulness (CF) labels. Tongyi Deep
Research produces no citations at all on \textsc{DeepResearchBench} and
\textsc{ResearchRubrics} (marked N/A), while LangChain and Smolagents show high
proportions of ``Unverifiable'' labels, indicating that cited sources
frequently cannot corroborate the associated claims.
Figure~\ref{fig:ci_scatter} visualizes the components of Citation
Integrity by plotting CF against Citation Attribution (CA). This scatter
plot reveals two distinct failure modes: LangChain attributes claims to
sources frequently (high CA, $\approx$ 75--80) but with poor
faithfulness (low CF, $\approx$ 10--20), meaning it cites often but
inaccurately. Conversely, Smolagents and Tongyi Deep Research rarely
attribute claims at all (low CA, $\approx$ 5--15), though when they do,
faithfulness is moderately higher (CF $\approx$ 35--55). Neither
strategy yields reliable citation behavior, underscoring that
open-source DRAs have yet to bridge the gap between source retrieval and
faithful evidence grounding.

\begin{figure*}
    \centering
    \includegraphics[width=\linewidth]{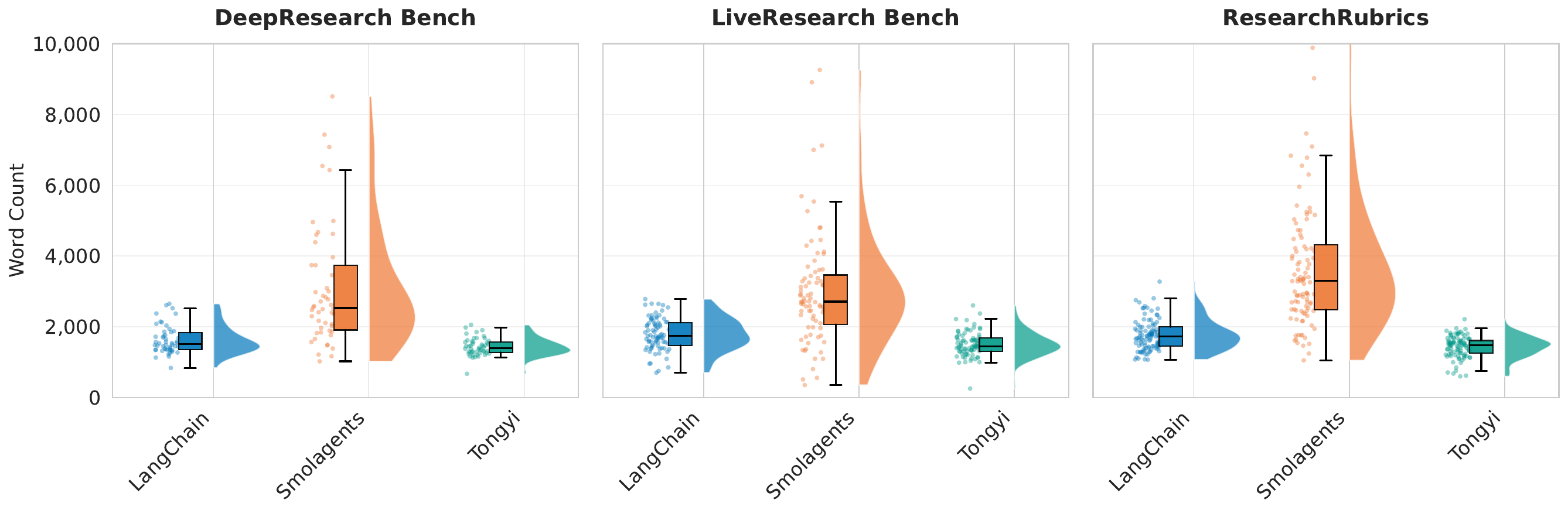}
    \caption{\textbf{Report length distributions across DRAs and datasets.} Word count distributions across DRAs and datasets. Smolagents Open Deep Research consistently exhibits the highest variance and mean length, while LangChain Open Deep Research and Tongyi Deep Research are more concise and consistent.}
    \label{fig:report_length_dist}
\end{figure*}

\paragraph{Report Length Analysis.} 
% Table~\ref{tab:report_lengths} and Figure~\ref{fig:report_length_dist} present the word count statistics ($Mean\pm Std$, Min, Max) and distribution, respectively, for each model. The standard deviations reveal significant differences in generation control. OpenAI Deep Research exhibits high instability, particularly on \textsc{LiveResearchBench}, where reports range from 2,916 to an extreme 29,800 words. In contrast, Gemini-3-Pro displays remarkable stability (Std $\approx$ 600 words), consistently producing focused, analyst-grade reports in the 4,000-word range.
Table~\ref{tab:report_lengths} and Figure~\ref{fig:report_length_dist}
present the word count statistics (Mean $\pm$ Std, Min, Max) and
distributions, respectively, for each agent. Smolagents Open DR
consistently generates the longest reports across all datasets, averaging
$\approx$ 3,000--3,700 words, with high variance and heavy-tailed
distributions (e.g., $\pm$ 2,066 words on \textsc{ResearchRubrics},
with outliers exceeding 14,500 words). LangChain Open DR and Tongyi Deep
Research produce more concise output in the 1,400--1,800 word range,
with Tongyi being the most consistent (Std $\approx$ 250--350 words).

\paragraph{Robustness Across Backbone Models. } We examined the impact of the underlying LLM backbone used to execute \AlgoName{}'s evaluation protocol on \textsc{DeepResearch Bench}. We compared the default model, Claude Sonnet 4.5, against DeepSeek-V3.2 and Kimi-K2.5 across our suite of static and adaptive metrics. The findings detailed in~\Cref{tab:judge_sensitivity} highlight \AlgoName{}'s robust evaluation signal. Even though absolute scores may change due to varying internal grading thresholds, the relative performance rankings of the evaluated deep research agents remain highly consistent. A minor exception is the Writing Quality (WQ) metric, for which Claude Sonnet 4.5 did not provide strong differentiation among agents, whereas DeepSeek-V3.2 and Kimi-K2.5 yield identical rankings.

\begin{table}[t]
\centering
\caption{\textbf{\AlgoName{} evaluation.} Static and adaptive metric scores for leading DRAs across three datasets.}
\begin{tabular}{ll|cccc|cc}
\toprule
& & \multicolumn{4}{c|}{\textbf{Static Metrics}} & \multicolumn{2}{c}{\textbf{Adaptive Metrics}} \\
\textbf{Dataset} & \textbf{DRA} & \textbf{WQ} & \textbf{Fact.} & \textbf{CI} & \textbf{DA} & \textbf{KIC} & \textbf{RQ} \\
\midrule
\multirow{3}{*}{DeepResearchBench} 
& LangChain Open DR    & 63.69 & 46.02 & \textbf{14.43} & \textbf{86.32} & 64.99 & 53.34 \\
& Smolagents Open DR   & 63.30 & 57.64 & 2.45  & 9.44  & \textbf{74.15} & \textbf{63.68} \\
& Tongyi Deep Research & \textbf{63.95} & \textbf{58.58} & 0.00  & 0.00  & 58.53 & 40.56 \\
\midrule
\multirow{3}{*}{LiveResearchBench}
& LangChain Open DR    & 60.75 & 38.11 & \textbf{11.90} & \textbf{84.00} & 63.66 & 59.04 \\
& Smolagents Open DR   & \textbf{61.71} & \textbf{56.90} & 9.67  & 24.41 & \textbf{73.55} & \textbf{68.82} \\
& Tongyi Deep Research & 60.03 & 51.69 & 2.92  & 5.71  & 55.17 & 46.95 \\
\midrule
\multirow{3}{*}{ResearchRubrics}
& LangChain Open DR    & 62.53 & 49.13 & \textbf{19.81} & \textbf{90.00} & 68.25 & 57.84 \\
& Smolagents Open DR   & \textbf{66.60} & \textbf{59.38} & 2.03  & 0.00  & \textbf{78.74} & \textbf{72.15} \\
& Tongyi Deep Research & 61.92 & 56.06 & 0.00  & 0.00  & 59.87 & 46.75 \\
\midrule[0.1em]
\multirow{3}{*}{\textbf{Aggregate}}
& LangChain Open DR    & 62.17 & 44.64 & \textbf{15.92} & \textbf{85.08} & 65.96 & 57.28 \\
& Smolagents Open DR   & \textbf{63.97} & \textbf{58.15} & 4.78  & 17.96 & \textbf{75.95} & \textbf{69.16} \\
& Tongyi Deep Research & 61.72 & 55.09 & 1.03  & 3.38  & 57.95 & 45.48 \\
\bottomrule
\end{tabular}
\label{tab:dra_results}
\end{table}


\clearpage

\begin{table*}
\centering
\caption{\textbf{Report length statistics across DRAs and datasets.} Word counts are reported as Mean $\pm$ Standard Deviation. Smolagents consistently produces the longest reports, while Tongyi Deep Research is the most concise.}
\begin{tabular}{llccc}
\toprule
\textbf{Dataset} & \textbf{DRA} & \textbf{Mean $\pm$ Std} & \textbf{Min} & \textbf{Max} \\
\midrule
\multirow{3}{*}{DeepResearchBench} 
 & LangChain Open DR    & $1628.3 \pm 398.3$  & 836  & 2650  \\
 & Smolagents Open DR          & $3092.0 \pm 1728.0$ & 1022 & 8511  \\
 & Tongyi Deep Research & $1431.0 \pm 248.7$  & 670  & 2055  \\
\midrule
\multirow{3}{*}{LiveResearchBench} 
 & LangChain Open DR    & $1792.0 \pm 459.9$  & 701  & 2784  \\
 & Smolagents Open DR          & $3009.3 \pm 1627.8$ & 352  & 9261  \\
 & Tongyi Deep Research & $1505.2 \pm 350.9$  & 254  & 2602  \\
\midrule
\multirow{3}{*}{ResearchRubrics} 
 & LangChain Open DR    & $1763.2 \pm 430.7$  & 1069 & 3275  \\
 & Smolagents Open DR          & $3735.2 \pm 2065.6$ & 1050 & 14554 \\
 & Tongyi Deep Research & $1436.1 \pm 301.4$  & 597  & 2214  \\
\bottomrule
\end{tabular}
\label{tab:report_lengths}
\end{table*}

\begin{table}[t]
\centering
\caption{\textbf{Backbone Model Comparison.} Metric scores on \textsc{DeepResearchBench} across three judge backbones. Agent rankings remain largely consistent regardless of the judge model used.}
% \small
\begin{tabular}{ll cccccc}
\toprule
& & \multicolumn{4}{c}{\textbf{Static Metrics}} & \multicolumn{2}{c}{\textbf{Adaptive Metrics}} \\
\cmidrule(lr){3-6} \cmidrule(lr){7-8}
\textbf{Judge Model} & \textbf{DRA} & \textbf{WQ} & \textbf{Fact.} & \textbf{CI} & \textbf{DA} & \textbf{KIC} & \textbf{RQ} \\
\midrule
\multirow{3}{*}{Claude Sonnet 4.5}
& LangChain Open DR    & 63.69 & 46.02 & \textbf{14.43} & \textbf{86.32} & 64.99 & 53.34 \\
& Smolagents Open DR   & 63.30 & 57.64 & 2.45  & 9.44  & \textbf{74.15} & \textbf{63.68} \\
& Tongyi Deep Research & \textbf{63.95} & \textbf{58.58} & 0.00  & 0.00  & 58.53 & 40.56 \\
\midrule
\multirow{3}{*}{DeepSeek V3.2}
& LangChain Open DR    & 66.19 & 47.35 & \textbf{17.24} & \textbf{86.14} & 63.86 & 67.68 \\
& Smolagents Open DR   & \textbf{67.10} & 54.45 & 6.02  & 60.72 & \textbf{73.57} & \textbf{71.89} \\
& Tongyi Deep Research & 65.27 & \textbf{57.46} & 0.00  & 6.22  & 57.00 & 56.20 \\
\midrule
\multirow{3}{*}{Kimi K2.5}
& LangChain Open DR    & 61.90 & 48.96 & \textbf{17.70} & \textbf{84.63} & 60.53 & 57.50 \\
& Smolagents Open DR   & \textbf{64.43} & 58.76 & 3.25  & 13.30 & \textbf{69.74} & \textbf{63.00} \\
& Tongyi Deep Research & 58.82 & \textbf{61.55} & 0.00  & 0.00  & 51.01 & 43.27 \\
\bottomrule
\end{tabular}
\label{tab:judge_sensitivity}
\end{table}

% \clearpage
% \section{Prompt Templates}\label{app:prompts}
% \subsection{Factuality}
% \input{prompts/factuality_content_extraction}
% \input{prompts/factuality_judgement}
% \subsection{Citation Integrity}
% \input{prompts/citatation_integrity_claim_extraction}
% \input{prompts/citation}

% \subsection{Protocol Creation}
% \input{prompts/protocol_creation_system_prompt}

\end{document}